\pdfoutput=1
\documentclass{article}

\usepackage{microtype}
\usepackage{graphicx}
\usepackage{subcaption}
\usepackage{booktabs} 

\usepackage{hyperref}

\usepackage{array}
\usepackage{amsmath}
\usepackage{booktabs}       
\usepackage{multirow}

\usepackage[abbreviations]{foreign}
\usepackage{tikz}
\usepackage{pgfplots}
\usepgfplotslibrary{fillbetween}
\usetikzlibrary{pgfplots.groupplots}
\usepgfplotslibrary{groupplots}

\usepackage{fp}
\usepackage{etoolbox,siunitx}
\usepackage{bm}
\usepackage{dsfont}
\usepackage{amsfonts}
\usepackage{mathtools}

\usepackage[accepted]{icml2021_arxiv}
\icmltitlerunning{SVMax: A Feature Embedding Regulairzer}

\begin{document}

\twocolumn[
\icmltitle{SVMax: A Feature Embedding Regularizer}




\begin{icmlauthorlist}
\icmlauthor{Ahmed Taha}{to}
\icmlauthor{Alex Hanson}{to}
\icmlauthor{Abhinav Shrivastava}{to}
\icmlauthor{Larry Davis}{to}
\end{icmlauthorlist}

\icmlaffiliation{to}{University of Maryland, College Park}


\icmlcorrespondingauthor{Ahmed Taha}{}
\icmlcorrespondingauthor{Alex  Hanson}{}

\icmlkeywords{Machine Learning, SVMax, Feature Embedding, Regularizer}

\vskip 0.3in
]



\printAffiliationsAndNotice{}  

\newcommand\sbullet[1][.5]{\mathbin{\vcenter{\hbox{\scalebox{#1}{$\bullet$}}}}}
\newcommand{\topic}[1]{\noindent\textbf{#1:}}

\newcommand{\recall}[1]{\FPeval{\result}{round(100.00*#1,2)}\result}

\newcommand{\margin}[1]{\FPeval{\result}{round((#1)*100.00,2)}\result}

\newcommand{\trim}[1]{\FPeval{\result}{round(#1,3)}\result}

\newcommand{\vanilla}{Vanilla}
\newcommand{\spread}{Spread-out}
\newcommand*{\rom}[1]{\uppercase\expandafter{\romannumeral #1\relax}}

\begin{abstract}
A neural network regularizer (\eg, weight decay) boosts performance by \textit{explicitly} penalizing the complexity of a network. In this paper, we penalize inferior network activations -- feature embeddings -- which in turn regularize the network's weights \textit{implicitly}. We propose singular value maximization (SVMax) to learn a more uniform feature embedding.  The SVMax regularizer supports both supervised and unsupervised learning. Our formulation mitigates model collapse and enables larger learning rates. We evaluate the SVMax regularizer using both retrieval and generative adversarial networks. We leverage a synthetic mixture of Gaussians dataset to evaluate SVMax in an unsupervised setting. For retrieval networks, SVMax achieves significant improvement margins across various ranking losses. Code available  at
\textit{https://bit.ly/3jNkgDt}	

\vspace{0.1in}




\end{abstract}

\section{Introduction}


A neural network's knowledge is embodied in \textit{both} its weights and activations. This difference manifests in how network pruning and knowledge distillation tackle the model compression problem. While pruning literature~\cite{li2016pruning,luo2017thinet,yu2018nisp} compresses models by removing less significant weights, knowledge distillation~\cite{hinton2015distilling} reduces computational complexity by matching a cumbersome network's last layer activations (logits).  This perspective, of weight-knowledge versus activation-knowledge, emphasizes how neural network literature is dominated by explicit weight regularizers. In contrast, this paper leverages singular value decomposition (SVD) to regularize a network through its last layer activations -- its feature embedding.

Our formulation is inspired by principal component analysis (PCA). Given a set of points and their covariance, PCA yields the set of orthogonal eigenvectors sorted by their eigenvalues. The principal component (first eigenvector) is the axis with the highest variation (largest eigenvalue) as shown in Figure~\ref{fig:PCA}. The eigenvalues from PCA, and similarly the singular values from SVD, provide insights about the feature embedding structure. As such, by regularizing the singular values, we reshape the feature embedding.



The \underline{main contribution} of this paper is to leverage the singular value decomposition of a network's activations to regularize the feature embedding. We achieve this objective through singular value maximization (SVMax). The SVMax regularizer is oblivious to both the input-class (labels) and the sampling strategy. Thus it promotes a uniform feature embedding in both supervised and unsupervised learning. Furthermore, we present a mathematical analysis of the mean singular value's lower and upper bounds. This analysis makes tuning the SVMax's balancing-hyperparameter easier, when the feature embedding is normalized to the unit circle.
 
 


SVMax promotes a uniform feature embedding. During training, SVMax speeds up convergence by enabling large learning rates. The SVMax regularizer integrates seamlessly with various ranking losses. We apply the SVMax regularizer to the last feature embedding layer, but the same formulation can be applied to intermediate layers. The SVMax regularizer mitigates model collapse in both retrieval networks and generative adversarial networks (GANs)~\cite{goodfellow2014generative,srivastava2017veegan,metz2016unrolled}. Furthermore, the SVMax regularizer is useful when training self/un-supervised feature embedding networks with a contrastive loss (\eg, CPC)~\cite{noroozi2017representation,oord2018representation,he2019momentum,tian2019contrastive}.


In summary, we propose singular value maximization to regularize the feature embedding. In addition, we present a mathematical analysis of the mean singular value's lower and upper bounds to reduce hyperparameter tuning (Sec.~\ref{sec:approach}). We quantitatively evaluate how SVMax significantly boosts the performance of ranking losses (Sec.~\ref{sec:retrieval}). And we provide a qualitative evaluation of using SVMax in the unsupervised learning setting via GAN training (Sec.~\ref{sec:gans}).
\input{intro_figure}
\section{Related Work}
Network weight regularizers dominate the deep learning regularizer literature because they support a large spectrum of tasks and architectures. Singular value decomposition (SVD) has been applied as a weight regularizer in several recent works ~\cite{zhang2018stabilizing,sedghi2018singular,guo2019regularization}. \citet{zhang2018stabilizing} employ SVD to avoid vanishing and exploding gradients in recurrent neural networks. Similarly,~\citet{guo2019regularization} bound the singular values of the convolutional layer around 1 to preserve the layer’s input and output norms. A bounded output norm mitigates the exploding/vanishing gradient problem. Weight regularizers share the common limitation that they do not enforce an explicit feature embedding objective and are thus ineffective against model collapse.


Feature embedding regularizers have also been extensively studied, especially for classification networks~\cite{rippel2015metric,wen2016discriminative,he2018triplet,hoffman2019robust,taha2020boosting}. These regularizers aim to maximize class margins, class compactness, or both simultaneously. For instance,~\citet{wen2016discriminative} propose center loss to explicitly learn class representatives and thus promote class compactness. In classification tasks, test samples are assumed to lie within the same classes of the training set,~\ie, closed-set identification. However, retrieval tasks, such as product re-identification, assume an open-set setting. Because of this, a retrieval network regularizer should aim to spread features across many dimensions to fully utilize the expressive power of the embedding space.

Recent literature~\cite{sablayrolles2018spreading,zhang2017learning} has recognized the importance of a spread-out feature embedding. However, this literature is tailored to triplet loss and therefore assumes a particular sampling procedure. In this paper, we leverage SVD as a regularizer because it is simple,  differentiable~\cite{ionescu2015training}, and class oblivious. SVD has been used to promote \textit{low} rank models to learn compact intermediate layer representations~\cite{kliegl2017trace,amartya2019learning}. This helps compress the network and speed up matrix multiplications on embedded devices (iPhone and Raspberry Pi). In contrast, we regularize the embedding space through a \textit{high} rank objective. By maximizing the mean singular value, we promote a higher rank representation -- a spread-out embedding.








\section{Singular Value Maximization (SVMax)}\label{sec:approach}
We first introduce our mathematical notation. Let $\mathcal{I}$ denote
the image space and $E_\mathcal{I} \in R^d$ denote the feature embeddings space, where $d$ is the dimension of the features. A feature embedding network is a function $F_{\theta}: \mathcal{I} \rightarrow E_\mathcal{I}$, parameterized by the network's weights $\theta$. We quantify similarity between an image pair $(\mathcal{I}_1, \mathcal{I}_2)$ via the Euclidean distance in feature space,~\ie, $\|E_{\mathcal{I}_1} - E_{\mathcal{I}_2} \|_2$.


During training, a $2D$ matrix $E \in R^{b\times d}$ stores $b$ samples' embeddings, where $b$ is the mini-batch size. Assuming $b\ge d$, the singular value decomposition (SVD) of $E$ provides the singular values $S =[s_1,.,s_i,.,s_d]$, where $s_1$ and $s_d$ are the largest and smallest singular values, respectively. We maximize the mean singular value,  $s_\mu = \frac{1}{d}\sum_{i=1}^{d}{s_i}$, to regularize the network's last layer activations -- the feature embedding. By maximizing the mean singular value, the deep network spreads out its embeddings. This has the added benefit of implicitly regularizing the network's weights $\theta$. The proposed SVMax regularizer integrates with both supervised and unsupervised feature embedding networks as follows
\begin{equation}\label{eq:svd_avg_s}
	L_{\text{NN}} = L_r - \lambda \frac{1}{d}\sum_{i=1}^{d}{s_i} = L_r - \lambda s_\mu,
\end{equation}
where $L_r$ is the original  loss and $\lambda$ is a  hyperparameter.



\textbf{Lower and Upper Bounds of the Mean Singular Value:} One caveat to equation~\ref{eq:svd_avg_s} is the hyperparameter $\lambda$. It is difficult to tune since the mean singular value $s_\mu$ depends on the range of values inside $E$ and its dimensions ($b,d$). Thus, changing the batch size or embedding dimension  requires a different $\lambda$. To address this, we constrain the embeddings to lie on the unit circle (L2-normalized) -- a common assumption in metric learning. This provides \textit{both} lower and upper bounds on ranking losses. This will also allow us to impose lower and upper bounds on $s_\mu$.

 
 
For an L2-normalized embedding $E$, the largest singular value $s_1$ is maximum when the matrix-rank of $E$ equals one,~\ie, $rank(E)=1$, and $s_i=0 \text{ for } i\in[2,d]$.~\citet{horn1991topics} provide an upper bound on this largest singular value $s_1$ as $s^\ast(E) \le \sqrt{||E||_1 ||E||_\infty}$. This holds in equality for all L2-normalized $E\in R^{b\times d}$ with $rank(E)= 1$. For an L2-normalized matrix $E$ with $||E||_1 =b$, and  $||E||_\infty=1$, this gives:
\begin{equation}
	s^\ast(E) = \sqrt{||E||_1 ||E||_\infty} = \sqrt{ b }.
\end{equation}
Thus, the lower bound $L$ on $s_\mu$ is $L = \frac{s^\ast(E)}{d} = \frac{\sqrt{ b }}{d}.$

Similarly, an upper bound is defined on the sum of the singular values~\cite{turkmen2007some,kong2018new,friedland2016computational}. This summation is formally known as the nuclear norm of a matrix $||E||_*$.~\citet{hu2015relations} established an upper bound on this summation using the Frobenius Norm $||E||_F$  as follows
\begin{equation}\label{eq:nuclear_norm}
||E||_* \le \sqrt{\frac{b \times d}{max(b,d)}} ||E||_F,\\
\end{equation}
where $||E||_F = { \left( \sum _{ i=1 }^{ rows }{ \sum _{ j=1 }^{ cols }{ { \left| { E }_{ ij } \right|  }^{ 2 } }  }  \right)  }^{ \frac { 1 }{ 2 }  } = \sqrt{ b }$ because of the L2-normalization assumption.

 
Accordingly, the lower and upper bounds of $s_\mu$ are $ [L,U] = [\frac{s^\ast(E)}{d} ,\frac{||E||_*}{d}]$. With these bounds, we rewrite our final loss function as follows
\begin{equation}\label{eq:svd_u_l}
L_{\text{NN}} = L_r + \lambda \exp\left(\frac{U-s_\mu}{U-L} \right).
\end{equation}
The SVMax regularizer  grows exponentially $\in [1, e]$. We employ this loss function in all our retrieval experiments. It is important to note that the L2-normalized assumption makes $\lambda$ tuning easier, but it is not required. Equation~\ref{eq:svd_u_l} makes the hyperparameter $\lambda$ only dependent on the range of $L_r$ which is also bounded for ranking losses. 



\textbf{Lower and Upper Bounds of Ranking Losses:} We briefly show that ranking losses are bounded when assuming an L2-normalized embedding. Equations~\ref{eq:svd_triplet} and~\ref{eq:svd_contrastive} show triplet and contrastive losses, respectively, and their corresponding bounds $[L,U]$.
\begin{align}\label{eq:svd_triplet}\nonumber
\text{TL}_{(a,p,n) \in T} &={ { \left[  { (D_{a,p}-D_{a,n} +m) }  \right]  }_{ + }  }\\
& \xrightarrow[]{[L,U]} \quad  [0,2+m], \\ \nonumber \label{eq:svd_contrastive}
\text{CL}_{(x,y)\in P}&= { { { \delta_{x,y}D_{x,y} } }+{(1 - \delta_{x,y})\left[ { m-D_{x,y} } \right]  }_{ + } } \\ & \xrightarrow[]{[L,U]} \quad  [0,2],
\end{align}
where ${ \left[ \sbullet[0.75] \right]  }_{ + }= max(0,\sbullet[0.75])$, $m < 2$ is the margin between classes, since $2$ is the maximum distance on the unit circle. $D_{x_1,x_2}=D(N(x_1),N(x_2))$; $N(\sbullet[0.75]) $ and $D(\sbullet[0.75], \sbullet[0.75])$ are the network's output-embedding and Euclidean distance, respectively. In equation~\ref{eq:svd_triplet}, $a$, $p$, and $n$ are the anchor, positive, and negative images in a single triplet $(a,p,n)$ from the triplets set $T$. In equation~\ref{eq:svd_contrastive}, $x$ and $y$ form a single pair of images from the pairs set $P$. $\delta_{x,y} = 1$ when $x$ and $y$ belong to the same class; zero otherwise. In the paper appendix, we (1) show similar analysis for N-pair and angular losses, and (2) provide an empirical SVMax evaluation on small training batches,~\ie,  $b < d$.






%
\section{Experiments}

In this section, we evaluate SVMax using both supervised and unsupervised learning. We leverage retrieval and generative adversarial networks for quantitative and qualitative evaluations, respectively.

\subsection{Retrieval Networks} \label{sec:retrieval}


\begin{table*}[t]
	\centering
	\scriptsize
	
	\caption{Quantitative evaluation on CUB-200-2011 with batch size $b=144$, embedding dimension $d=128$ and multiple learning rates $lr=\{0.01,0.001,0.0001\}$. $\triangle_{R@1}$ column indicates the R@1 improvement margin relative to the vanilla ranking loss. A large learning rate $lr$ increases the chance of model collapse, while a small $lr$ slows convergence. $\lambda$ is dependent on the ranking loss.}
		\setlength{\tabcolsep}{3pt}
	\begin{tabular}{@{} l cccc c cccc c cccc   @{}}
		\toprule
		& \multicolumn{4}{c}{$lr=0.01$} && \multicolumn{4}{c}{$lr=0.001$} && \multicolumn{4}{c}{$lr=0.0001$}\\
 		\cmidrule{2-5} \cmidrule{7-10} \cmidrule{12-15}
 				Method & NMI & R@1 & R@8 & $\triangle_{R@1}$ && NMI & R@1 & R@8 & $\triangle_{R@1}$ && NMI & R@1 & R@8 & $\triangle_{R@1}$\\
 				\midrule
 				
		& \multicolumn{14}{c}{Contrastive} \\
 				\cmidrule{2-15}
		
		Vanilla &       0.435      &  \recall{0.257258609}   & \recall{0.5887913572}   & - &
				  &        0.443      & \recall{0.2867994598}    & \recall{0.6470290344}  &  - &
				   &        0.413     &  \recall{0.2449358542}   &  \recall{0.5953747468}  &  -      \\

		Spread-out &      0.440       &  \recall{0.2454422687}   & \recall{0.5715732613}       & \margin{0.2454422687-0.257258609} &
		& 0.479         &  \recall{0.3212356516}    & \recall{0.668298447}         &    \margin{0.3212356516-0.2867994598} &
		 & \bf0.458       &   \bf\recall{0.3185347738}   &  \bf\recall{0.6745442269}    & \margin{0.3185347738-0.2449358542}   \\

	SVMax (Ours) & \bf0.527       &   \bf\recall{0.4125590817}   & \bf\recall{0.7523632681}    & \margin{0.4125590817-0.257258609} &
	 & \bf0.547       &    \bf \recall{0.4311276165}    & \bf\recall{0.7726198515}     & \margin{0.4311276165-0.2867994598}  &
	&      0.449       &     \recall{0.2955773126}   &  \recall{0.6549628629}    &   \margin{0.2955773126-0.2449358542}   \\
		
		 				\midrule
				& \multicolumn{14}{c}{Triplet Loss} \\
		\cmidrule{2-15}
				Vanilla &       0.496     & \recall{0.2933828494}    &   \recall{0.6796083727}  & - &
		& 0.477      &   \recall{0.2888251182}  &  \recall{0.6460162053}   &   - &
		& 0.449     & \recall{0.2486495611}    & \recall{0.6114112086}    &  -   \\

				Spread-out  & 0.545          &    \recall{0.4360229575}    &   \recall{0.7697501688}     &  \margin{0.4360229575-0.2933828494}&
		&    \bf0.557         &   \bf\recall{0.440243079}  &  \bf\recall{0.7854490209}    & \margin{0.440243079-0.2888251182} &
&    0.435         & \bf\recall{0.2832545577}    & \recall{0.6433153275}     & \margin{0.2832545577-0.2486495611}  \\
		
		SVMax $\lambda = 1$ (Ours) &  \bf0.556      &   \recall{0.4321404456}    &   \recall{0.7743079001}   &  \margin{0.4321404456-0.2933828494}&
		&       0.527      &     \recall{0.3912896691}  &  \recall{0.7417285618}    & \margin{0.3912896691-0.2888251182} &
		& 0.401        &   \recall{0.2506752194}    & \recall{0.6001012829}  & \margin{0.2506752194-0.2486495611} \\

		SVMax $\lambda = 0.1$ (Ours) &       0.547      &   \bf\recall{0.4380486158}    &   \bf\recall{0.7797096556}     &  \margin{0.4380486158-0.2933828494} &
&      \bf 0.557      &     \recall{0.4388926401}  &  \recall{0.7844361918}      & \margin{0.4388926401-0.2888251182} &
&    \bf0.436     &   \recall{0.2822417286}    & \bf\recall{0.6439905469}       & \margin{0.2822417286-0.2486495611} \\

 				\midrule
				& \multicolumn{14}{c}{N-pair} \\
\cmidrule{2-15}

		Vanilla &       0.402    & \recall{0.1895678596}   & \recall{0.5032072924}    & -  &
		&        0.452      &   \recall{0.2765023633}  &   \recall{0.6309925726}  &    -&
		& 0.455       &   \recall{0.3141458474}  &  \recall{0.669480081}     & -  \\
		
		Spread-out &     0.416        &   \recall{0.2064483457}   & \recall{0.528021607}            &  \margin{0.2064483457-0.1895678596}&
		&        0.483     &   \recall{0.3246117488}  &   \recall{0.6640783255}  &    \margin{0.3246117488-0.2765023633} &
		& 		0.474         &     \recall{0.3338960162}  &  \recall{0.6880486158}  &      \margin{0.3338960162-0.3141458474} \\
		
		SVMax (Ours) &        \bf0.483     &  \bf\recall{0.3462187711}   & \bf\recall{0.6811276165}   &   \margin{0.3462187711-0.1895678596} &
		&       \bf0.547      &    \bf\recall{0.4378798109}  &   \bf\recall{0.773126266	}  &       \margin{0.4378798109-0.2765023633} &
		&  \bf0.488        &     \bf\recall{0.3413234301}  &  \bf\recall{0.6991897367}  & \margin{0.3413234301-0.3141458474}  \\

 				\midrule
				& \multicolumn{14}{c}{Angular} \\
\cmidrule{2-15}

Vanilla &             0.470      &  \recall{0.2854490209}   &	\recall{0.6002700878}    &  -  &
&     0.508        &  \recall{0.3894328157}   &	\recall{0.7282241729}   & - &
&    \bf0.538        &  \recall{0.4179608373}   &	\recall{0.7618163403}& -    \\

Spread-out &       .471      &  \recall{0.282916948}   &   \recall{0.6026333558}  &  \margin{0.282916948-0.2854490209} &
&     0.508      &  \recall{0.3896016205}   &   \recall{0.7285617826}  &   \margin{0.3896016205-0.3894328157} &
&     \bf0.538        &  \recall{0.4181296421}   &	\recall{0.7623227549}& \margin{0.4181296421-0.4179608373}    \\


SVMax (Ours) & \bf0.487    &   \bf\recall{0.3288318704}   &  \bf\recall{0.6627278866}  &   \margin{0.3288318704-0.2854490209} &
&      \bf0.523       &   \bf\recall{0.4128966914}   &  \bf\recall{0.7471303174}  &   \margin{0.4128966914-0.3894328157} &
&      0.531	       &  \bf \recall{0.4199864956}   &	\bf\recall{0.7629979743}&  \margin{0.4199864956-0.4179608373}    \\
		
		\bottomrule
	\end{tabular}

	\label{tbl:quan_cub}
	
\end{table*}


\begin{figure*}[t]
	\centering
	\scriptsize
	\begin{tikzpicture}
	\begin{groupplot}[group style = {group size = 4 by 1, horizontal sep = 20pt}, 
	height=3.7cm,
	symbolic x coords={0.0001,0.001,0.01},
	xtick=data,
	x label style={at={(axis description cs:0.5,-0.05)},anchor=north}
	]
	\nextgroupplot[title=Angular,ylabel=R@1,		y label style={at={(axis description cs:0.15,.5)}},
	legend style = { legend columns = -1, legend to name = grouplegend,}]
	\addplot[
	color=blue,
	dashed,
	]
	coordinates {
		(0.0001,37.01881687) 
		(0.001,62.45234289) 
		(0.01,32.78809495) }; 
	\addlegendentry{\vanilla}

	\addplot[
color=red,
dashed,
mark=*,
]
	coordinates {
	(0.0001,36.94502521) 
	(0.001,62.13257902) 
	(0.01,31.74271307) 
}; \addlegendentry{\spread}

	\addplot[
color=red,
solid,
]
	coordinates {
	(0.0001,35.12483089) 
	(0.001,63.15336367) 
	(0.01,38.58074038) 
}; \addlegendentry{SVMax}

		\nextgroupplot[title=N-pair,	legend style = { legend columns = -1, legend to name = grouplegend,}]
	\addplot[
	color=blue,
	dashed,
	]
	coordinates {
		(0.0001,24.48653302) 
		(0.001,25.98696347) 
		(0.01,23.84700529) 
	}; \addlegendentry{\vanilla}
	
	\addplot[
	color=red,
	dashed,
mark=*,
	]
	coordinates {
		(0.0001,26.71258148) 
		(0.001,35.59217808) 
		(0.01,29.95941459) 
	}; \addlegendentry{\spread}
	
	\addplot[
	color=red,
	solid,
	]
	coordinates {
		(0.0001,25.5442135) 
		(0.001,48.91157299) 
		(0.01,42.98364285) 
	}; \addlegendentry{SVMax}	
	
			\nextgroupplot[title=Triplet,	legend style = { legend columns = -1, legend to name = grouplegend,}]
	\addplot[
	color=blue,
	dashed,
	]
	coordinates {
		(0.0001,16.6646169) 
		(0.001,12.21251999) 
		(0.01,8.387652195) 
	}; \addlegendentry{\vanilla}
	
	\addplot[
	color=red,
	dashed,
mark=*,
	]
	coordinates {
		(0.0001,22.10060263) 
		(0.001,35.75206002) 
		(0.01,62.20637068) 
	}; \addlegendentry{\spread}
	
	\addplot[
	color=red,
	solid,
	]
	coordinates {
		(0.0001,20.55097774) 
		(0.001,27.70876891) 
		(0.01,63.27634977) 
	}; \addlegendentry{SVMax}	

	\nextgroupplot[title=Contrastive,	legend style = { legend columns = -1, legend to name = grouplegend,}]
	\addplot[
	color=blue,
	dashed,
	]
	coordinates {
		(0.0001,17.53781823) 
		(0.001,23.82240807) 
		(0.01,27.8194564) 
	}; \addlegendentry{\vanilla}
	
	\addplot[
	color=red,
	dashed,
mark=*,
	]
	coordinates {
		(0.0001,24.95388021) 
		(0.001,31.70581724) 
		(0.01,33.66129627) 
	}; \addlegendentry{\spread}
	
	\addplot[
	color=red,
	solid,
	]
	coordinates {
		(0.0001,20.12052638) 
		(0.001,36.96962243) 
		(0.01,61.3946624) 
	}; \addlegendentry{SVMax}

	\end{groupplot}
	\node[below] at ($(group c2r1.south) +(1.8,-0.5)$) {\pgfplotslegendfromname{grouplegend}}; 
	\end{tikzpicture}
	
	\caption{Quantitative evaluation on Stanford CARS196. X and Y-axis denote the learning rate $lr$ and recall@1 performance, respectively.}
	\label{fig:quan_cars}
\end{figure*}
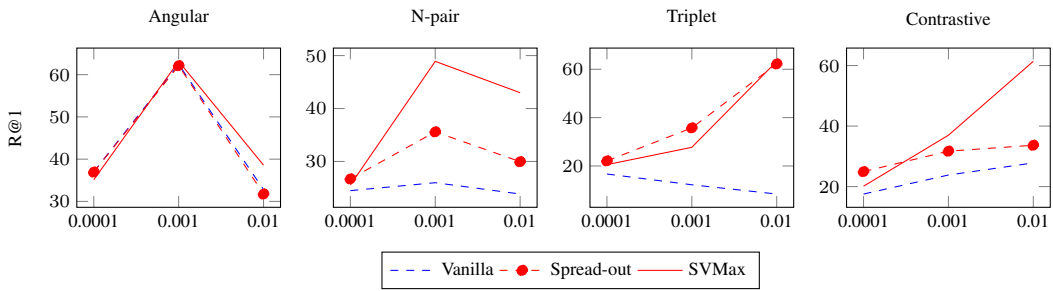

\topic{Technical Details} We evaluate SVMax quantitatively using three datasets: CUB-200-2011~\cite{wah2011caltech}, Stanford CARS196~\cite{krause20133d}, and Stanford Online Products~\cite{oh2016deep}. We use GoogLeNet~\cite{szegedy2015going} and ResNet50~\cite{he2016deep}; both pretrained on ImageNet~\cite{deng2009imagenet} and fine-tuned for $K$ iterations. These are standard retrieval datasets and architectures. By default, the embedding $\in R^{d=128}$ is normalized to the unit circle. In all experiments, a batch size $b=144$ is employed, the learning rate $lr$ is fixed for $K/2$ iterations then decayed polynomially to $1e-7$ at iteration $K$. We use the SGD optimizer with $0.9$ momentum. Each batch contains $p$ different classes and $l$ different samples per class. For example, triplet loss employs $p=24$ different classes and $l=6$ instances per class. The mini-batch of N-pair loss contains $72$ classes and a single positive pair per class,~\ie, $p=72$ and $l=2$. This same mini-batch setting is used for angular loss. For contrastive loss, $p=36$ and $l=4$ are divided into $72$ positive and $72$ negative pairs. For both CUB-200 and CARS196, $K=5,000$ iterations; for Stanford Online Products, $K=20,000$.

\topic{Baselines} We evaluate SVMax using contrastive~\cite{hadsell2006dimensionality}, hard triplet~\cite{hoffer2015deep,hermans2017defense}, N-pair~\cite{sohn2016improved} and angular~\cite{wang2017deep} losses. We use the margin $m = 1$ for contrastive loss, $m = 0.2$ for triplet loss, and the angle bound $\alpha=45^{\circ}$ for angular loss. Similar to SVMax, multiple regularizers~\cite{kumar2016learning,zhang2017learning,amartya2019learning,chen2019energy} promote a uniform embedding space. Unlike SVMax, these regularizers require a supervised setting to push anchor-negative pairs apart. We employ the spread-out regularizer~\cite{zhang2017learning} as a baseline for its simplicity, with default hyperparameter $\alpha=1$. To enable the spread-out regularizer on non-triplet ranking losses, we pair every anchor with a random negative sample from the training mini-batch. 




\topic{Evaluation Metrics} For quantitative evaluation, we use the Recall@K metric and Normalized Mutual Info (NMI) on the test split.

\topic{The hyperparameter} $\lambda=1$ for both contrastive and N-pair losses, $\lambda=0.1$ for triplet loss, and $\lambda=2$ for angular loss. We fix $\lambda$ across datasets, architectures, and other hyperparameters ($b,d$).

\begin{table*}[t]
	\centering
	\scriptsize
	
	\caption{Quantitative evaluation on Stanford Online Products.}
		\setlength{\tabcolsep}{3pt}
	\begin{tabular}{@{} l cccc c cccc c cccc   @{}}
		\toprule
		& \multicolumn{4}{c}{$lr=0.01$} && \multicolumn{4}{c}{$lr=0.001$} && \multicolumn{4}{c}{$lr=0.0001$}\\
 		\cmidrule{2-5} \cmidrule{7-10} \cmidrule{12-15}
 				Method & NMI & R@1 & R@8 & $\triangle_{R@1}$ && NMI & R@1 & R@8 & $\triangle_{R@1}$ && NMI & R@1 & R@8 & $\triangle_{R@1}$\\
 				\midrule
 				
		& \multicolumn{14}{c}{Contrastive} \\
 				\cmidrule{2-15}
		
		Vanilla &       \trim{0.816}      &  \recall{0.1823080229}   & \recall{0.3407325378}   & - &
				  &        \trim{0.820}      & \recall{0.2869657201}    & \recall{0.432663383}  &  - &
				   &        \trim{0.813}     &  \recall{0.3429969257}   &  \recall{0.4849095898}  &  -      \\

		Spread-out &      \trim{0.811}       &  \recall{0.1887045056}   & \recall{0.3574096724}       & \margin{0.1887045056-0.1823080229} &
		& \trim{0.822}         &  \recall{0.2996595154}    & \recall{0.4669267132}         &    \margin{0.2996595154-0.2869657201} &
		 & \trim{0.824}       &   \recall{0.3615252388}   &  \recall{0.5121814155}    & \margin{0.3615252388-0.3429969257}   \\

	SVMax (Ours) & \bf\trim{0.875}       &   \bf\recall{0.6181613831}   & \bf\recall{0.7890482959}    & \margin{0.6181613831-0.1823080229} &
	 & \bf\trim{0.854}       &    \bf \recall{0.5394367128}    & \bf\recall{0.7092327526}     & \margin{0.5394367128-0.2869657201}  &
	&      \bf\trim{0.832}       &     \bf\recall{0.4195894351}   &  \bf\recall{0.5743942349}    &   \margin{0.4195894351-0.3429969257}   \\
		
		 				\midrule
				& \multicolumn{14}{c}{Triplet Loss} \\
		\cmidrule{2-15}
				Vanilla &       \bf\trim{0.891}     & \bf\recall{0.7196456315}    &   \bf\recall{0.8624012429}  & - &
		& \bf\trim{0.873} 	&   \recall{0.6408548478}  &  \recall{0.8007173317}   &   - &
		& \bf\trim{0.840}     & \recall{0.4629433738}    & \recall{0.6256652673}    &  -   \\

				Spread-out & \trim{0.890}          &    \recall{0.7160259165}    &   \recall{0.8572774454}     &  \margin{0.7160259165-0.7196456315}&
		&    \trim{0.872}        &   \bf\recall{0.64225976}  &  \recall{0.8009817857}    & \margin{0.64225976-0.6408548478} &
&    \bf\trim{0.840}         & \bf\recall{0.4667944861}    & \bf\recall{0.6303923837}     & \margin{0.4667944861-0.4629433738}  \\
		
		SVMax $\lambda = 1$ (Ours) &  \trim{0.868}      &   \recall{0.6381937787}    &   \recall{0.809493901}   &  \margin{0.6381937787-0.7196456315}&
		&      \trim{0.8565490354}      &     \recall{0.5804436217}  &  \recall{0.751396648}    & \margin{0.5804436217-0.6408548478} &
		& \trim{0.8361983502}        &   \recall{0.4462331824}    & \recall{0.6076493339}  & \margin{0.4462331824-0.4629433738} \\

		SVMax $\lambda = 0.1$ (Ours) &       \trim{0.888807646}      &   \recall{0.7147697597}    &   \recall{0.8597236455}     &  \margin{0.7147697597-0.7196456315} &
&       \trim{0.871825566}      &     \bf\recall{0.6422928168}  & \bf \recall{0.801428052}      & \margin{0.6422928168-0.6408548478} &
&    \bf\trim{0.8401235929}     &   \recall{0.466397805}    & \recall{0.6294998512}       & \margin{0.466397805-0.4629433738} \\

 				\midrule
				& \multicolumn{14}{c}{N-pair} \\
\cmidrule{2-15}

		Vanilla &       \trim{0.7982232547}    & \recall{0.128574262}   & \recall{0.2453472612}    & -  &
		&        \trim{0.8145504918}      &   \recall{0.2383061717}  &   \recall{0.389689597}  &    -&
		& \trim{0.8184928555}       &   \recall{0.3398400053}  &  \recall{0.4856368385}     & -  \\
		
		Spread-out &     \trim{0.8029755479}        &   \recall{0.1657631153}   & \recall{0.3191464745}            &  \margin{0.1657631153-0.128574262}&
		&        \trim{0.824}     &   \recall{0.3288155763}  &   \recall{0.5034379029}  &    \margin{0.3288155763-0.2383061717} &
		& 		\trim{0.8247887217}         &     \recall{0.3738554097}  &  \recall{0.5255363459}  &      \margin{0.3738554097-0.3398400053} \\
		
		SVMax (Ours) &        \bf\trim{0.8706153862}     &  \bf\recall{0.5775676837}   & \bf\recall{0.7605203134}   &   \margin{0.5775676837-0.128574262} &
		&       \bf\trim{0.8575928702}      &    \bf\recall{0.5469901821}  &   \bf\recall{0.7156788205}  &       \margin{0.5469901821-0.2383061717} &
		&  \bf\trim{0.8352363215}        &     \bf\recall{0.4304155235}  &  \bf\recall{0.5877656937}  & \margin{0.4304155235-0.3398400053}  \\

 				\midrule
				& \multicolumn{14}{c}{Angular} \\
\cmidrule{2-15}

Vanilla &             \trim{0.8834716217}      &  \recall{0.6283263363}   &	\recall{0.8012958249}    &  -  &
&     \bf\trim{0.8846105172}        &  \recall{0.66928366}   &	\recall{0.8211794651}   & - &
&    \bf\trim{0.8558839013}        &  \recall{0.5428746157}   &	\recall{0.7113814419}& -    \\

Spread-out &       \trim{0.8831741191}      &  \recall{0.6273181052}   &   \recall{0.7996099303}  &  \margin{0.6273181052-0.6283263363} &
&     \bf\trim{0.884811205}      &  \recall{0.6691018479}   &   \recall{0.8208654259}  &   \margin{0.6691018479-0.66928366} &
&     \bf\trim{0.8562706193}        &  \recall{0.5430398995}   &	\recall{0.7110178176}& \margin{0.5430398995-0.5428746157}    \\


SVMax (Ours) & \bf\trim{0.8849977151}    &   \bf\recall{0.6543915904}   &  \bf\recall{0.817295296}  &   \margin{0.6543915904-0.6283263363} &
&      \trim{0.8838251843}       &   \bf\recall{0.6727876764}   &  \bf\recall{0.8247495951}  &    \margin{0.6727876764-0.66928366} &
&      \trim{0.8552658179}	       &  \bf \recall{0.5488083039}   &	\bf \recall{0.7147367029}&  \margin{0.5488083039-0.5428746157}    \\
		
		\bottomrule
	\end{tabular}

	\label{tbl:quan_stanford}
	
\end{table*}

\topic{Results} Tables~\ref{tbl:quan_cub} and~\ref{tbl:quan_stanford} present quantitative retrieval evaluation on CUB-200 and Stanford Online Products datasets -- both using GoogLeNet. These tables provide in depth analysis and emphasize our improvement margins on a small and large dataset. Figure~\ref{fig:quan_cars} provides quantitative evaluation on Stanford CARS196. We report quantitative evaluation on ResNet50 in the paper appendix. Our training hyperparameters -- learning rate $lr$ and number of iterations $K$ -- do not favor a particular ranking loss in these experiments.




We evaluate SVMax on various learning rates. A large learning rate, \eg, $lr=0.01$, speeds up convergence, but increases the chance of model collapse. In contrast, a small rate, \eg, $lr=0.0001$, is likely to avoid model collapse but is slow to converge. This undesirable effect is tolerable for small datasets -- where increasing the number of training iterations $K$ does not drastically increase the overall training time -- but it is infeasible for large datasets.  For contrastive and N-pair losses, SVMax significantly outperforms both the vanilla and spread-out baselines with larger learning rates. A small $lr$ slows convergence and all approaches are roughly equivalent. The spread-out regularizer~\cite{zhang2017learning} and its hyperparameters are tuned for triplet loss. Thus, for this particular ranking loss, the SVMax and spread-out regularizers are on par.


In our experiments, we employ a large learning rate because it is the simplest factor to induce model collapse. However, the learning rate is not the only factor. Another factor is the training dataset size and its intra-class variations.  A small dataset with large intra-class variations increases the chances of a model collapse. For example, a pair of dissimilar birds from the same class justifies a model collapse when coupled with a large learning rate. The hard triplet loss experiments emphasize this point because every anchor is paired with the hardest positive and negative samples. On small fine-grained datasets like CUB-200 or CARS196, the vanilla hard triplet loss suffers significantly. Yet, the same implementation is superior on a big dataset like Stanford Online Products. By carefully tuning the training hyperparameter on CUB-200, it is possible to avoid a degenerate solution. However, this tedious tuning process is unnecessary when using either the spread-out or the SVMax regularizer.



The  vanilla N-pair loss underperforms because it does not support feature embedding on the unit circle. Both spread-out and SVMax mitigate this limitation. For angular loss, a bigger $\lambda=2$ is employed to cope with the angular loss range. SVMax is a class oblivious regularizer. Thus, $\lambda$ should be significant enough to contribute to the loss function without dominating the ranking loss.




\citet{wu2017sampling} show that the distance between any anchor-negative pair, which is randomly sampled from an $n$-dimensional unit sphere, follows the normal distribution $N(\sqrt{2},\frac{1}{2n})$. This mean distance $\sqrt{2}$ is large relative to the triplet loss margin $m=0.2$, but comparable to the contrastive loss margin $m=1$. Accordingly, triplet loss converges to zero after a few iterations, because most triplets satisfy the margin $m=0.2$ constraint. When triplet loss equals zero, the SVMax regularizer with $\lambda=1$ becomes the dominant term. However, SVMax should not dominate because it is oblivious to data annotations; it equally pushes anchor-positive and anchor-negative pairs apart. Reducing $\lambda$ to $0.1$ solves this problem.

A less aggressive triplet loss~\cite{schroff2015facenet,xuan2020improved} is another way to avoid model collapse. For instance,~\citet{schroff2015facenet} have proposed a triplet loss variant that employs semi-hard negatives. The semi-hard triplet loss is more stable than the aggressive hard triplet and lifted structured losses~\cite{oh2016deep}. Unfortunately, the semi-hard triplet loss assumes a large mini-batch ($b=1,800$ in \citet{schroff2015facenet}), which is impractical. Furthermore, when model collapse is avoided, aggressive triplet loss variants achieve superior performance~\cite{hermans2017defense}. In contrast, SVMax only requires a larger mini-batch than the embedding dimension,~\ie, $b \ge d$, a natural constraint for retrieval networks which favor compact embedding dimensions. Additionally, SVMax makes no assumptions about the sampling procedure. Thus, unlike~\cite{sablayrolles2018spreading,zhang2017learning}, SVMax supports various supervised ranking losses.

\input{svmax_sota}

  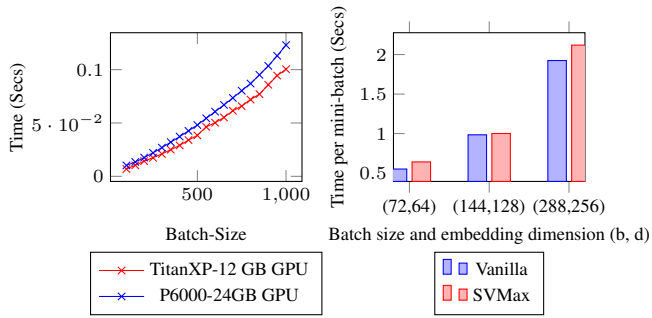
\begin{figure}[t]
	\centering
	\scriptsize
	\begin{tikzpicture}
		\begin{axis}[
			xlabel=Batch-Size,
			width=0.5\linewidth,
			legend style={at={(0.5,-0.9)},anchor=south,legend columns=1},
			ylabel=Time (Secs)]
			\addplot[color=red,mark=x] coordinates {
				(100,0.00683133602142334)(150,0.01033165454864502)(200,0.014278340339660644)(250,0.017412376403808594)(300,0.021005702018737794)(350,0.024968218803405762)(400,0.028973150253295898)(450,0.03399226665496826)(500,0.03836610317230225)(550,0.04636082649230957)(600,0.05023343563079834)(650,0.0552018404006958)(700,0.06150376796722412)(750,0.06576361656188964)(800,0.07193944454193116)(850,0.07702713012695313)(900,0.08586366176605224)(950,0.09444360733032227)(1000,0.1003941535949707)
			};
			\addlegendentry{TitanXP-12 GB GPU}
			
			\addplot[color=blue,mark=x] coordinates {
				(100,0.01021122932434082)(150,0.013435769081115722)(200,0.017537522315979003)(250,0.021992969512939452)(300,0.02689979076385498)(350,0.031996417045593264)(400,0.03724756240844727)(450,0.042590785026550296)(500,0.04807350635528564)(550,0.054406356811523435)(600,0.06063675880432129)(650,0.06694846153259278)(700,0.0735081672668457)(750,0.08026692867279053)(800,0.08698348999023438)(850,0.0951530933380127)(900,0.10352113246917724)(950,0.11296710968017579)(1000,0.12301249504089355)
			};
			\addlegendentry{P6000-24GB GPU}
			
		\end{axis}
	\end{tikzpicture}\begin{tikzpicture}
		\begin{axis}[
			ylabel=Time per mini-batch (Secs),
			xlabel=\text{Batch size and embedding dimension (b, d)},
			width=0.5\linewidth,
			legend style={at={(0.5,-0.9)},anchor=south,legend columns=1},
			y label style={at={(axis description cs:0.2,.5)}},
			ybar,
			bar width=7pt,
			xtick=data,
			symbolic x coords={
				\text{(72,64)},\text{(144,128)},\text{(288,256)}},
			]
			
			\addplot 
			coordinates {
				(\text{(72,64)},0.5516372251510621) (\text{(144,128)},0.9845351457595826)(\text{(288,256)},1.9245308542251587)
			};
			
			\addplot 
			coordinates {
				(\text{(72,64)},0.6414985656738281) (\text{(144,128)},1.0014208984375)(\text{(288,256)},2.1194742202758787)
			};

			
			

			\legend{Vanilla,SVMax}
		\end{axis}
	\end{tikzpicture}
	\caption{(Left) Timing analysis for the Tensorflow (TF) \texttt{tf.linalg.svd} function. The x-axis denotes the batch size $b$, and the y-axis denotes the running time in seconds. We time this TF function using two different GPUs. (Right) Timing analysis for a mini-batch training time using MobileNet. The x-axis denotes both the batch size $b$ and the embedding dimension $d$. The y-axis denotes the batch training time in seconds.} 
	\label{fig:svmax_analysis}
\end{figure}

 \topic{SVMax's Computational Complexity} We compute the singular values  $S =[s_1,.,s_i,.,s_d]$ using TensorFlow (TF) \texttt{tf.linalg.svd}. This function runs on the GPU. We did not notice any computational overhead or numerical instability during training. Figure~\ref{fig:svmax_analysis} (Left) provides a timing analysis of the TF function using square matrices. For a typical mini-batch size ($b<256$), the function takes around $0.01$ seconds. This speed depends on the GPU specification and recent GPUs would perform faster. Figure~\ref{fig:svmax_analysis} (Right) provides a timing analysis for the mini-batch training time using MobileNet. SVMax adds minimal overhead compared to the overhead of performing gradient descent on a deep network. We conclude that for a reasonable batch size $b$ and embedding dimension $d$, SVMax adds minimal computational complexity to the training process.
 
 

\newcommand{\VaraintsImgSize}{0.08}
\begin{figure}[t]
	\centering
	\scriptsize
	\setlength\tabcolsep{1.0pt} 
	\renewcommand{\arraystretch}{0.0}	
	\begin{tabular}{@{}>{\centering}m{0.35in}
			>{\centering}m{\VaraintsImgSize\textwidth} 
			>{\centering}m{\VaraintsImgSize\textwidth} |
			>{\centering}m{\VaraintsImgSize\textwidth} 
			>{\centering\arraybackslash}m{\VaraintsImgSize\textwidth}
		}
		\toprule
		 &Vanilla GAN & Vanilla GAN + SVMax & Unrolled GAN \\(5 steps) & Unrolled GAN  + SVMax\\
		\midrule
		Step 1 & \includegraphics[width=\VaraintsImgSize\textwidth,height=\VaraintsImgSize\textwidth]{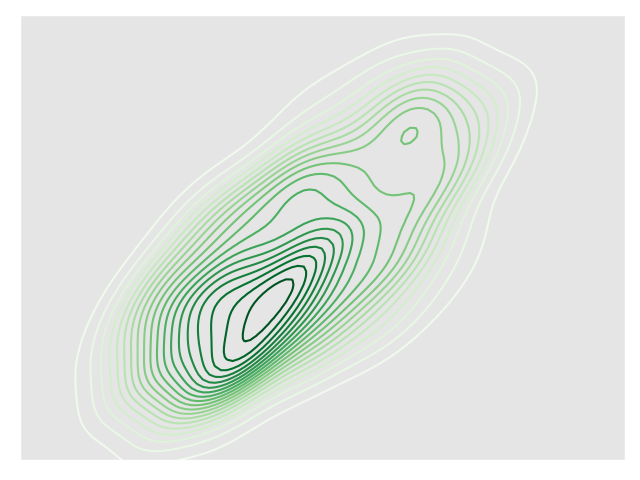} & \includegraphics[width=\VaraintsImgSize\textwidth,height=\VaraintsImgSize\textwidth]{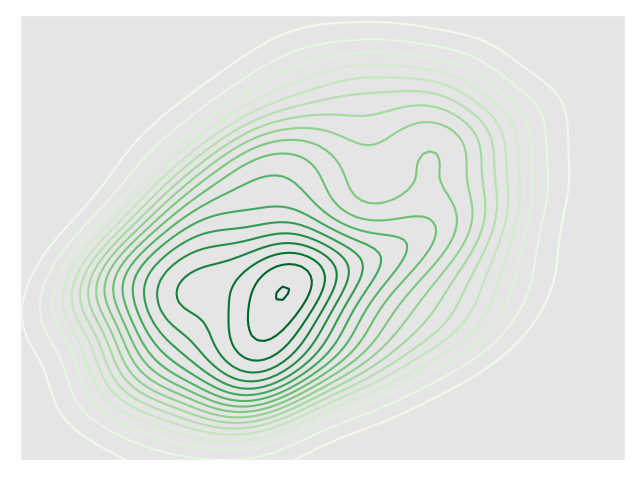} & \includegraphics[width=\VaraintsImgSize\textwidth,height=\VaraintsImgSize\textwidth]{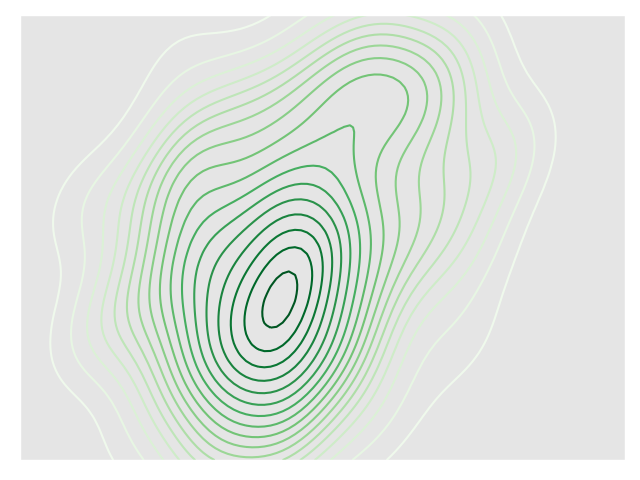} & \includegraphics[width=\VaraintsImgSize\textwidth,height=\VaraintsImgSize\textwidth]{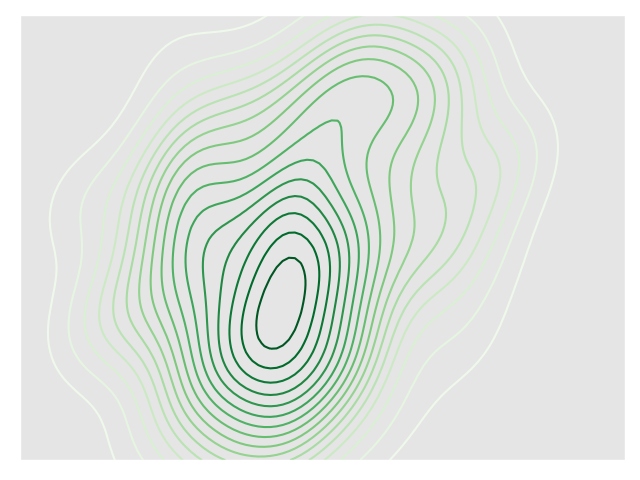} \\
		
		Step 5k& \includegraphics[width=\VaraintsImgSize\textwidth,height=\VaraintsImgSize\textwidth]{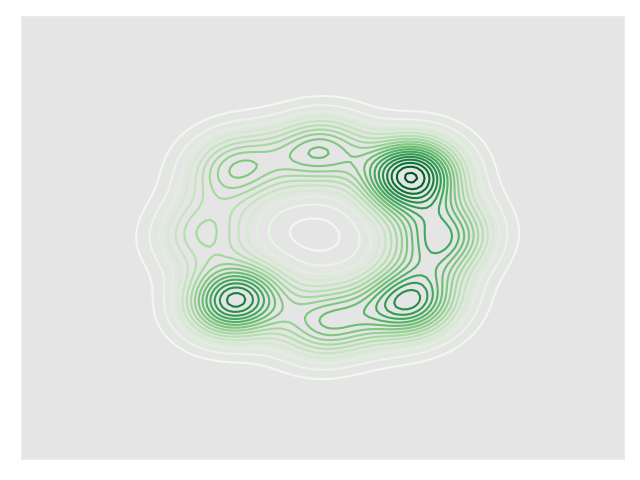} & \includegraphics[width=\VaraintsImgSize\textwidth,height=\VaraintsImgSize\textwidth]{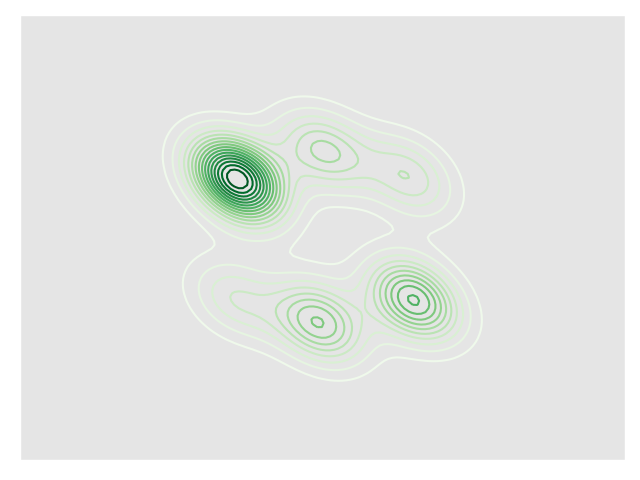} & \includegraphics[width=\VaraintsImgSize\textwidth,height=\VaraintsImgSize\textwidth]{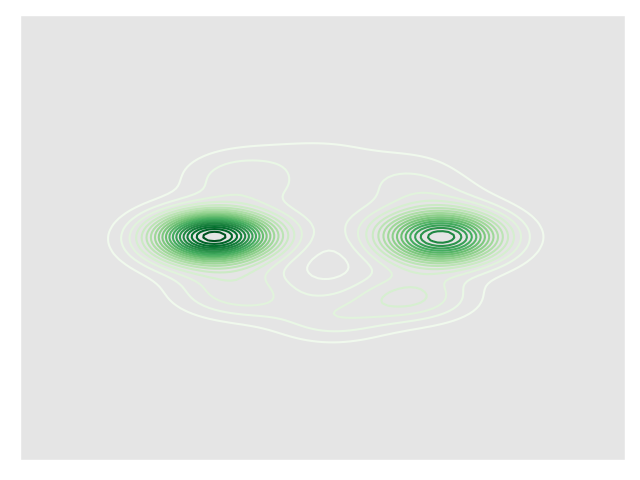} & \includegraphics[width=\VaraintsImgSize\textwidth,height=\VaraintsImgSize\textwidth]{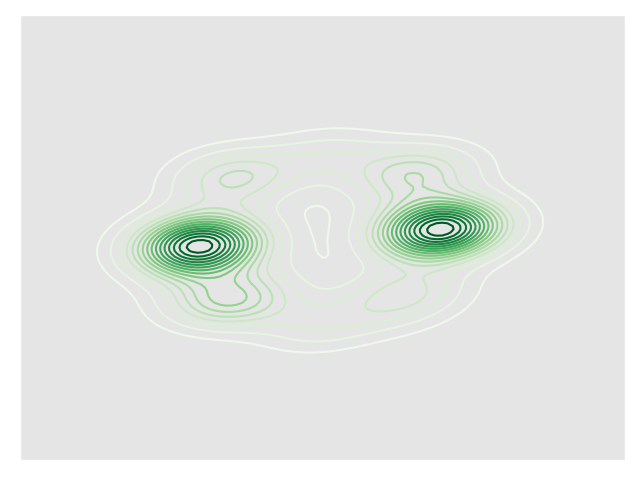} \\
		
		Step 10k &\includegraphics[width=\VaraintsImgSize\textwidth,height=\VaraintsImgSize\textwidth]{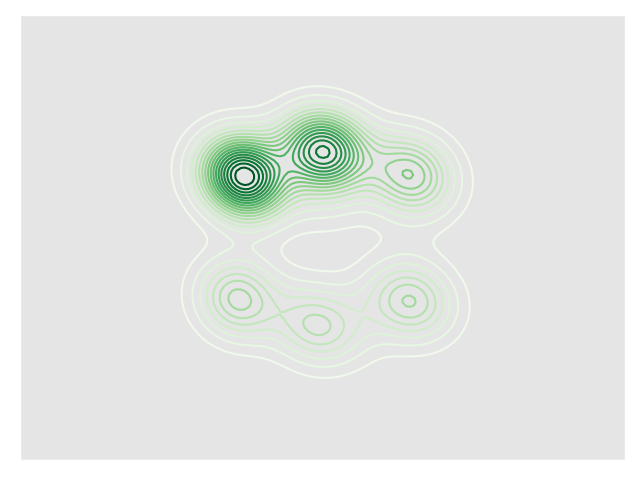}  & \includegraphics[width=\VaraintsImgSize\textwidth,height=\VaraintsImgSize\textwidth]{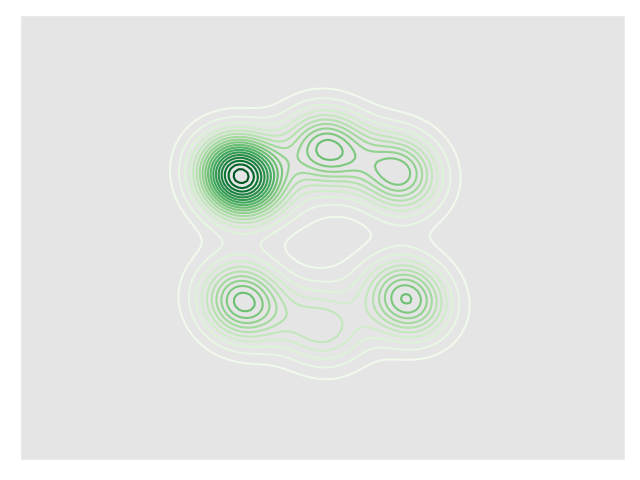} & \includegraphics[width=\VaraintsImgSize\textwidth,height=\VaraintsImgSize\textwidth]{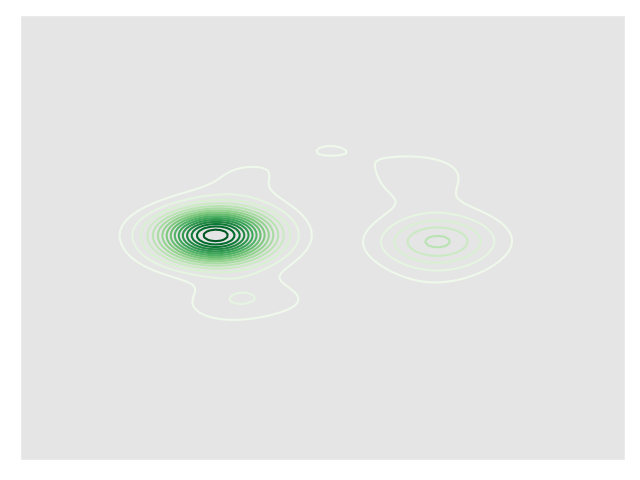} & \includegraphics[width=\VaraintsImgSize\textwidth,height=\VaraintsImgSize\textwidth]{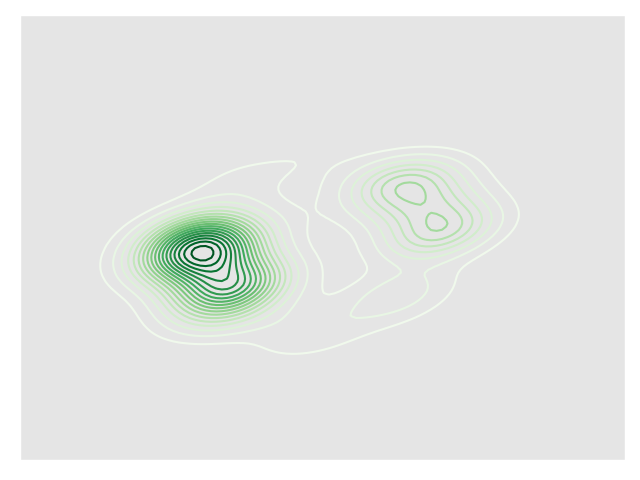} \\
		
		Step 15k & \includegraphics[width=\VaraintsImgSize\textwidth,height=\VaraintsImgSize\textwidth]{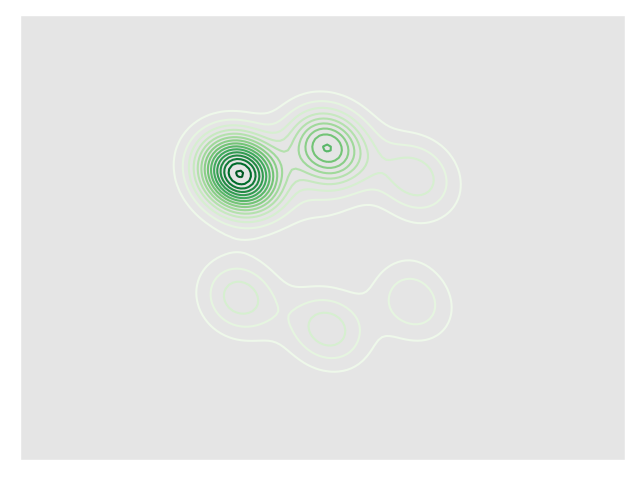} & \includegraphics[width=\VaraintsImgSize\textwidth,height=\VaraintsImgSize\textwidth]{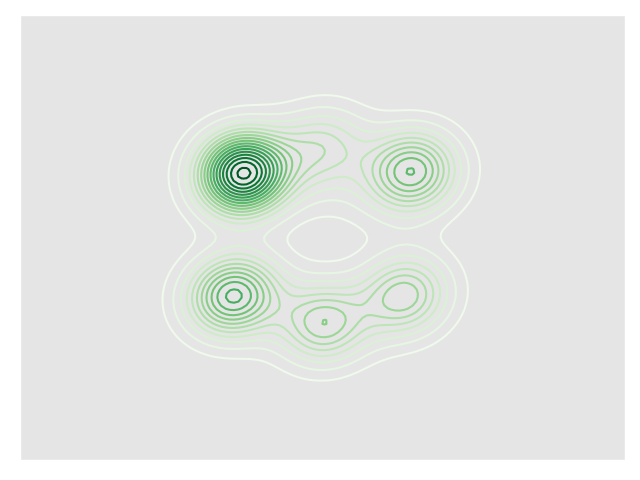} & \includegraphics[width=\VaraintsImgSize\textwidth,height=\VaraintsImgSize\textwidth]{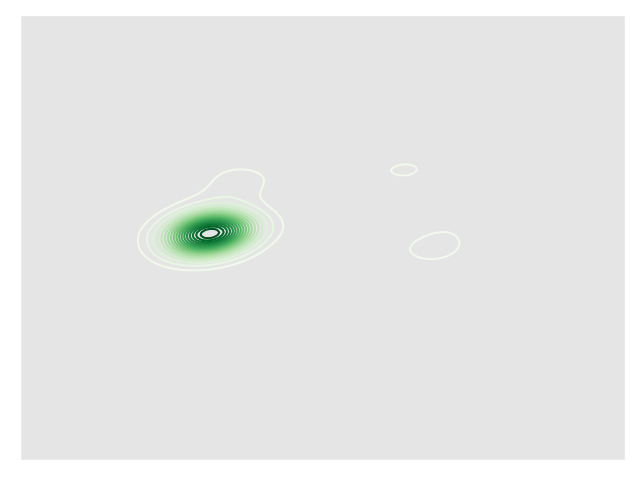} & \includegraphics[width=\VaraintsImgSize\textwidth,height=\VaraintsImgSize\textwidth]{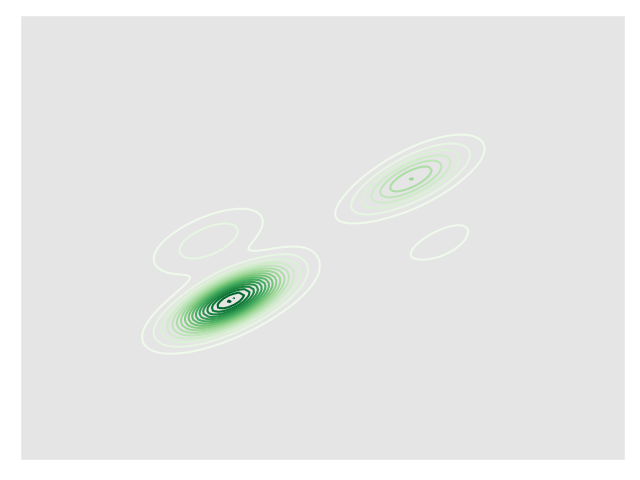} \\
		
		Step 20k & \includegraphics[width=\VaraintsImgSize\textwidth,height=\VaraintsImgSize\textwidth]{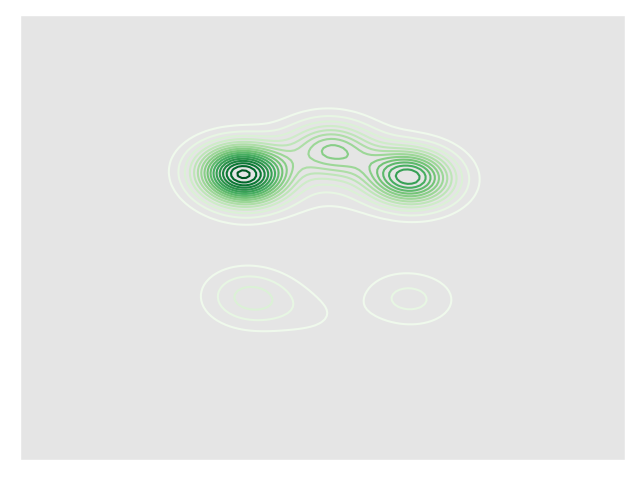} & \includegraphics[width=\VaraintsImgSize\textwidth,height=\VaraintsImgSize\textwidth]{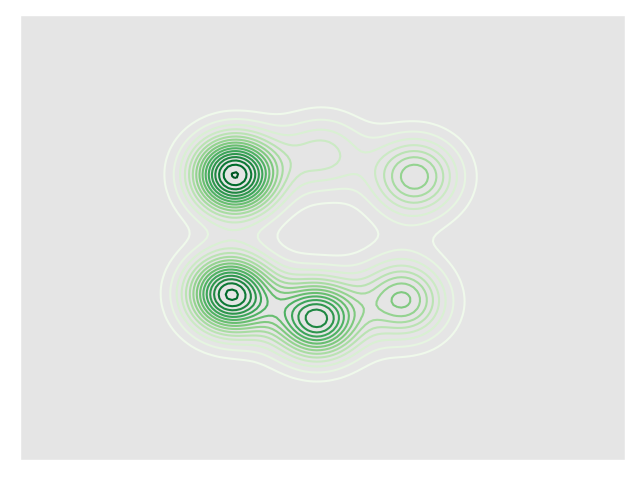} & \includegraphics[width=\VaraintsImgSize\textwidth,height=\VaraintsImgSize\textwidth]{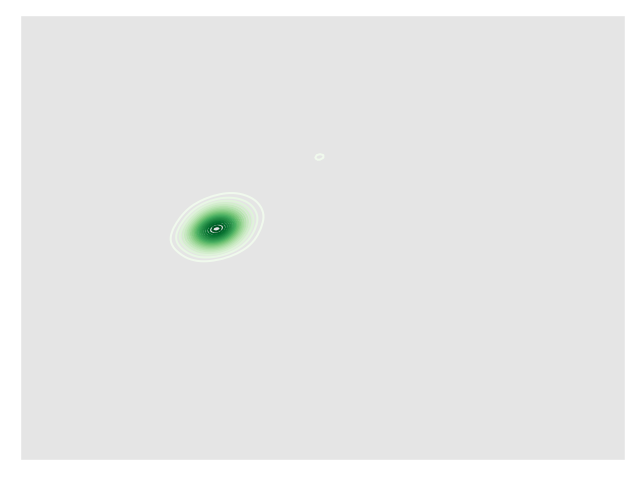}  & \includegraphics[width=\VaraintsImgSize\textwidth,height=\VaraintsImgSize\textwidth]{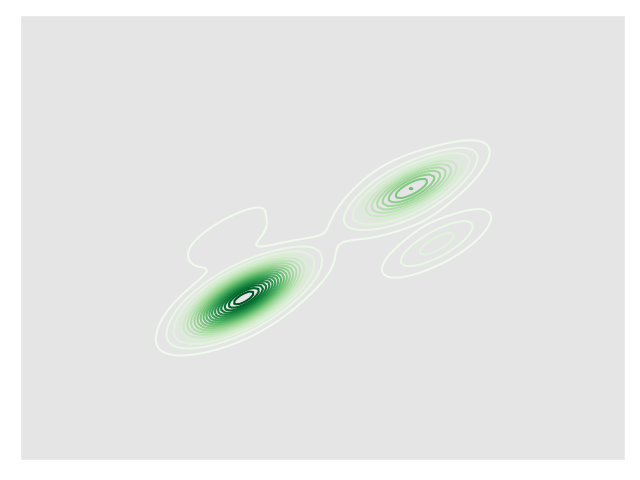}  \\
		
		Step 25k & \includegraphics[width=\VaraintsImgSize\textwidth,height=\VaraintsImgSize\textwidth]{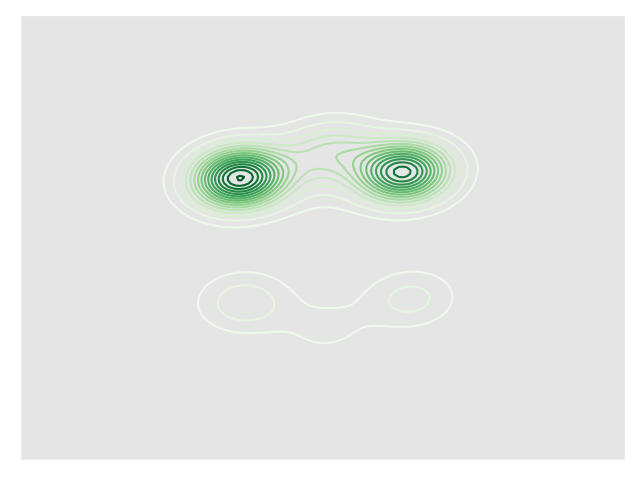} & \includegraphics[width=\VaraintsImgSize\textwidth,height=\VaraintsImgSize\textwidth]{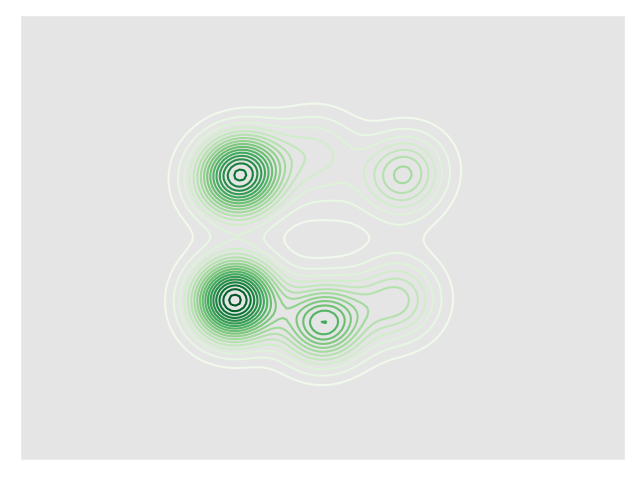}  & \includegraphics[width=\VaraintsImgSize\textwidth,height=\VaraintsImgSize\textwidth]{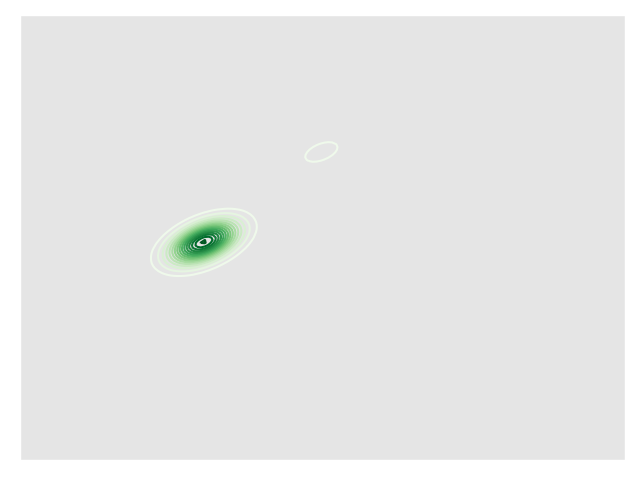} & \includegraphics[width=\VaraintsImgSize\textwidth,height=\VaraintsImgSize\textwidth]{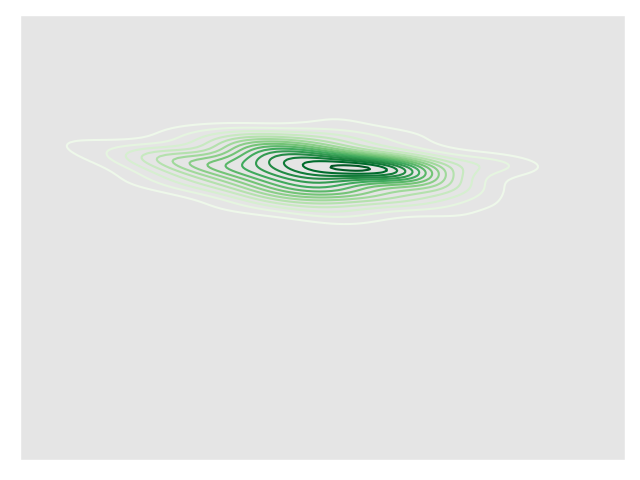} \\
		\midrule
		Target & \includegraphics[width=\VaraintsImgSize\textwidth,height=\VaraintsImgSize\textwidth]{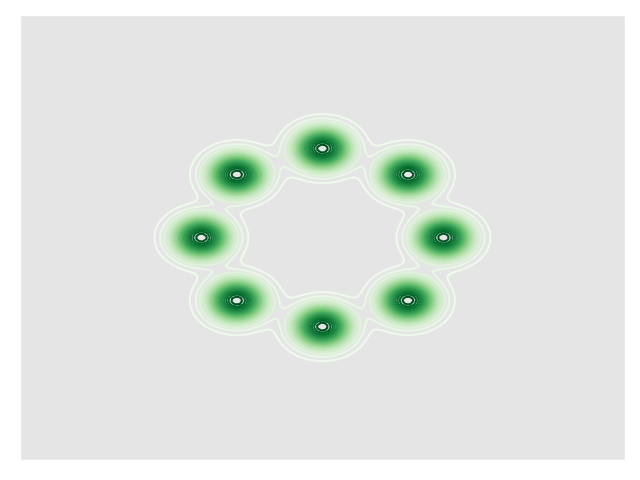}& \includegraphics[width=\VaraintsImgSize\textwidth,height=\VaraintsImgSize\textwidth]{figures/gans_r2/target} &
		\includegraphics[width=\VaraintsImgSize\textwidth,height=\VaraintsImgSize\textwidth]{figures/gans_r2/target}&
		\includegraphics[width=\VaraintsImgSize\textwidth,height=\VaraintsImgSize\textwidth]{figures/gans_r2/target}
		\\
		\bottomrule
	\end{tabular}

	\caption{SVMax mitigates model collapse in a GAN trained on a toy 2D mixture of Gaussians dataset. 
		Rows show heatmaps of the generator distributions at different training steps. The final row shows the groundtruth distribution. The first column shows the distributions generated by training a vanilla GAN suffering a model collapse. 
		The second column shows the generated distribution when penalizing the generator's fake embedding with SVMax. 
		The third and fourth columns show two distributions generated using an unrolled-GAN  without and with SVMax, respectively. 
		This high resolution figure is best viewed on a screen with zoom capabilities.
	}
	
	\label{fig:qual_gan_one_colunm}
\end{figure}
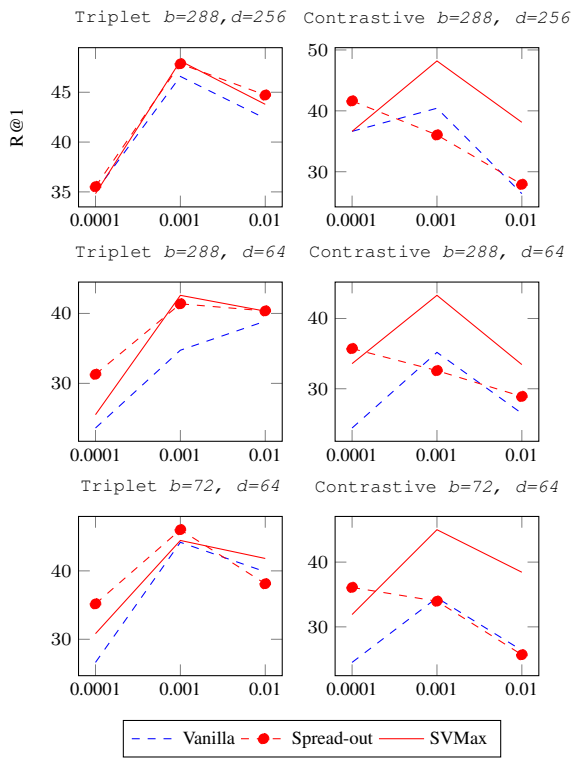
\begin{figure}[t]
	\centering
	\scriptsize
	\begin{tikzpicture}
	\begin{groupplot}[group style = {group size = 2 by 3, horizontal sep = 20pt}, 
	height=3.7cm,
	symbolic x coords={0.0001,0.001,0.01},
	xtick=data,
	x label style={at={(axis description cs:0.5,-0.05)},anchor=north}
	]
	\nextgroupplot[title=\texttt{Triplet \textit{b=288,d=256}},ylabel=R@1,		y label style={at={(axis description cs:0.15,.5)}},
	legend style = { legend columns = -1, legend to name = grouplegend,}]
	
	\addplot[	color=blue,	dashed,	]
	coordinates {
		(0.0001,35.02700878) 
		(0.001,46.5901418) 
		(0.01,42.37002026) 
	}; \addlegendentry{\vanilla}
	\addplot[color=red,dashed,mark=*,]
	coordinates {
	(0.0001,35.53342336) 
	(0.001,47.85617826) 
	(0.01,44.71640783) 
}; \addlegendentry{\spread}

	\addplot[color=red,solid,]
	coordinates {
	(0.0001,34.82444294) 
	(0.001,48.14314652) 
	(0.01,43.78798109) 
}; \addlegendentry{\approach{}}	

			\nextgroupplot[title=\texttt{Contrastive \textit{b=288, d=256}},	legend style = { legend columns = -1, legend to name = grouplegend,}]
\addplot[	color=blue,	dashed,	]
coordinates {
	(0.0001,36.63065496) 
	(0.001,40.42876435) 
	(0.01,26.40108035) 
}; \addlegendentry{\vanilla}

\addplot[	color=red,	dashed,	mark=*,]
coordinates {
	(0.0001,41.62727887) 
	(0.001,36.03983795) 
	(0.01,27.92032411) 
}; \addlegendentry{\spread}

\addplot[	color=red,	solid,]
coordinates {
	(0.0001,36.63065496) 
	(0.001,48.19378798) 
	(0.01,38.11613774) 
}; \addlegendentry{\approach{}}

		\nextgroupplot[title=\texttt{Triplet \textit{b=288, d=64}},	legend style = { legend columns = -1, legend to name = grouplegend,}]
	\addplot[
	color=blue,	dashed,	]
	coordinates {
		(0.0001,23.61580014) 
		(0.001,34.74004051) 
		(0.01,38.89264011) 
	}; \addlegendentry{\vanilla}
	
	\addplot[	color=red,	dashed,	mark=*,]
	coordinates {
		(0.0001,31.29642134) 
		(0.001,41.37407157) 
		(0.01,40.3612424) 
	}; \addlegendentry{\spread}
	
	\addplot[	color=red,	solid,	]
	coordinates {
		(0.0001,25.52329507) 
		(0.001,42.58946658) 
		(0.01,40.32748143) 
	}; \addlegendentry{SVMax}	

	\nextgroupplot[title=\texttt{Contrastive \textit{b=288, d=64}},	legend style = { legend columns = -1, legend to name = grouplegend,}]
\addplot[	color=blue,	dashed,	]
coordinates {
	(0.0001,24.40918298) 
	(0.001,35.17893315) 
	(0.01,26.48548278) 
}; \addlegendentry{\vanilla}

\addplot[	color=red,	dashed,	mark=*,]
coordinates {
	(0.0001,35.71910871) 
	(0.001,32.57933828) 
	(0.01,28.8993923) 
}; \addlegendentry{\spread}

\addplot[	color=red,	solid,]
coordinates {
	(0.0001,33.57528697) 
	(0.001,43.298447) 
	(0.01,33.44024308) 
}; \addlegendentry{SVMax}

		\nextgroupplot[title=\texttt{Triplet \textit{b=72, d=64}},	legend style = { legend columns = -1, legend to name = grouplegend,}]
	\addplot[
	color=blue,	dashed,	]
	coordinates {
		(0.0001,26.62052667) 
		(0.001,44.17623228) 
		(0.01,39.87170831) 
	}; \addlegendentry{\vanilla}
	
	\addplot[	color=red,	dashed,	mark=*,]
	coordinates {
		(0.0001,35.17893315) 
		(0.001,46.01620527) 
		(0.01,38.11613774) 
	}; \addlegendentry{\spread}
	
	\addplot[	color=red,	solid,	]
	coordinates {
		(0.0001,30.82376772) 
		(0.001,44.44632005) 
		(0.01,41.81296421) 
	}; \addlegendentry{SVMax}

		\nextgroupplot[title=\texttt{Contrastive \textit{b=72, d=64}},	legend style = { legend columns = -1, legend to name = grouplegend,}]
	\addplot[	color=blue,	dashed,	]
	coordinates {
		(0.0001,24.5104659) 
		(0.001,34.48683322) 
		(0.01,26.36731938) 
	}; \addlegendentry{\vanilla}
	
	\addplot[	color=red,	dashed,	mark=*,]
	coordinates {
		(0.0001,36.09047941) 
		(0.001,33.96353815) 
		(0.01,25.69209993) 
	}; \addlegendentry{\spread}
	
	\addplot[	color=red,	solid,]
	coordinates {
		(0.0001,31.90411884) 
		(0.001,45.0033761) 
		(0.01,38.43686698) 
	}; \addlegendentry{SVMax}	
	
	\end{groupplot}
	\node[below] at ($(group c1r3.south) +(1.75,-0.5)$) {\pgfplotslegendfromname{grouplegend}}; 
	\end{tikzpicture}
	
	\caption{Quantitative evaluation on CUB-200-2011 with various batch sizes $b=\{288,72\}$ and embedding dimensions $d=\{256,64\}$ to demonstrate the stability of our hyperparameter. $\lambda=1$ for contrastive loss and  $\lambda=0.1$ for triplet loss.}
	\label{fig:quan_cub_mobile}
	\vspace{-0.1in}
\end{figure}
\newcommand{\svdAblation}{0.12}
\begin{figure*}[t]
	\centering
	\scriptsize
	\setlength\tabcolsep{1.0pt} 
	\renewcommand{\arraystretch}{0.0}	
	
	\begin{tabular}{@{}>{\centering}m{0.5in} >{\centering}m{\svdAblation\textwidth}  >{\centering}m{\svdAblation\textwidth} >{\centering}m{\svdAblation\textwidth} >{\centering}m{\svdAblation\textwidth} >{\centering}m{\svdAblation\textwidth} >{\centering}m{\svdAblation\textwidth} >{\centering\arraybackslash}m{\svdAblation\textwidth}}
		\toprule
		Method &Epoch 1 & 		Epoch 2 & 		Epoch 4 & Epoch 8 & Epoch 16 & Epoch 32 & Epoch 64 \\
		\midrule
		Contrastive &\includegraphics[width=\svdAblation\textwidth,height=\svdAblation\textwidth]{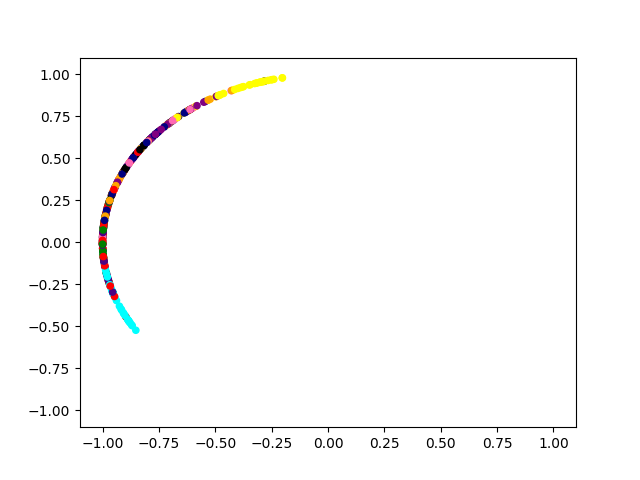} & \includegraphics[width=\svdAblation\textwidth,height=\svdAblation\textwidth]{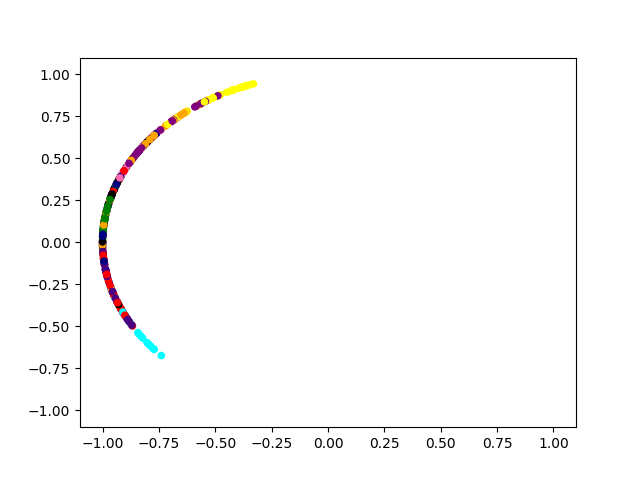} & \includegraphics[width=\svdAblation\textwidth,height=\svdAblation\textwidth]{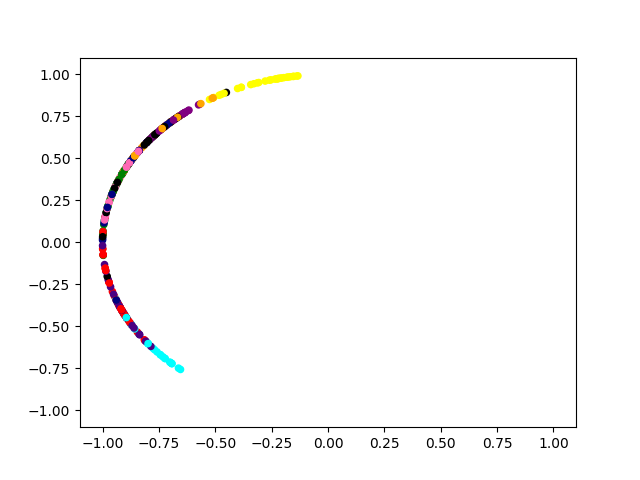} &
		\includegraphics[width=\svdAblation\textwidth,height=\svdAblation\textwidth]{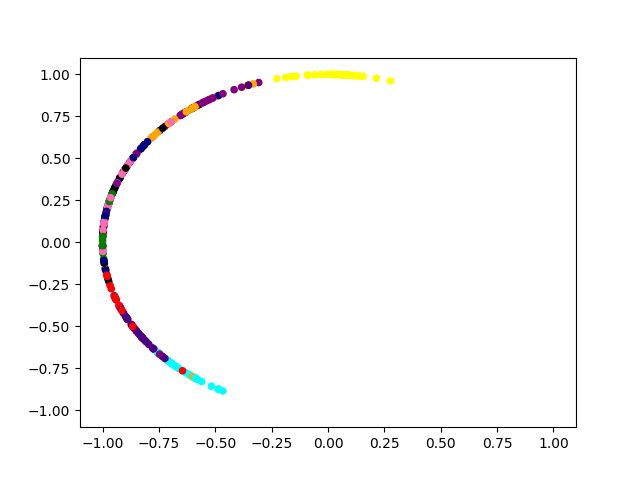} &
		\includegraphics[width=\svdAblation\textwidth,height=\svdAblation\textwidth]{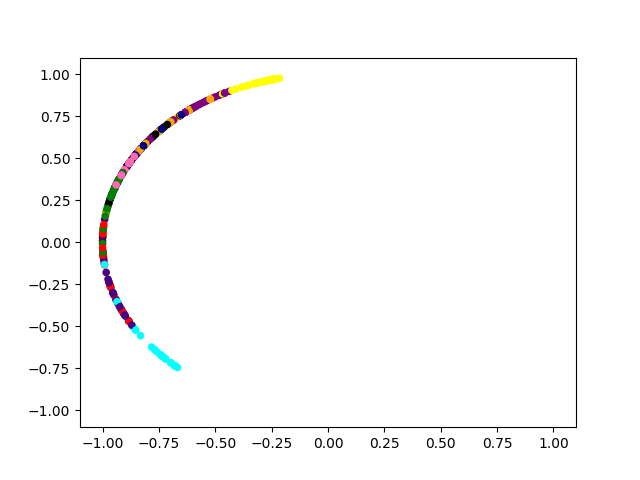} &
		\includegraphics[width=\svdAblation\textwidth,height=\svdAblation\textwidth]{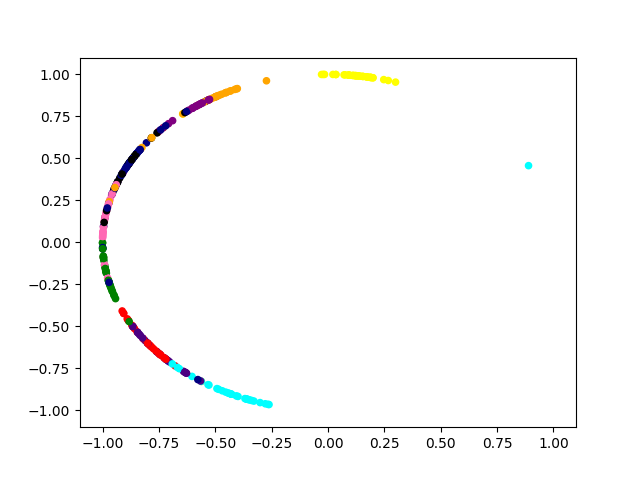} &
		\includegraphics[width=\svdAblation\textwidth,height=\svdAblation\textwidth]{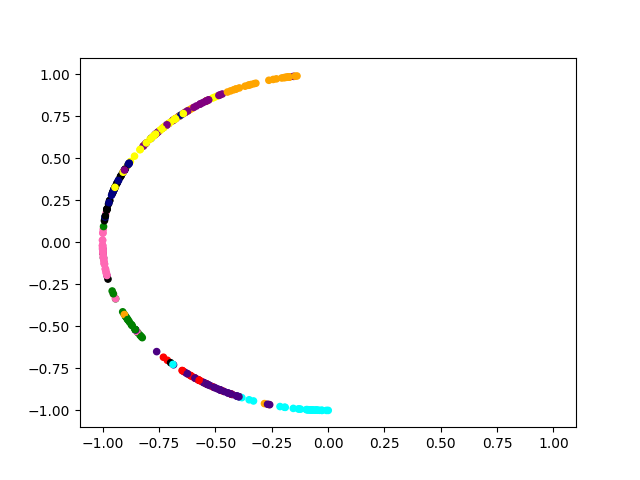} 
		\\
		
		Contrastive + SVMax &\includegraphics[width=\svdAblation\textwidth,height=\svdAblation\textwidth]{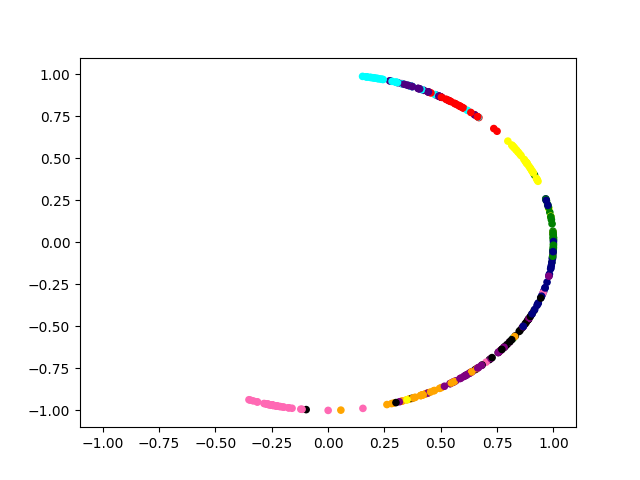} & \includegraphics[width=\svdAblation\textwidth,height=\svdAblation\textwidth]{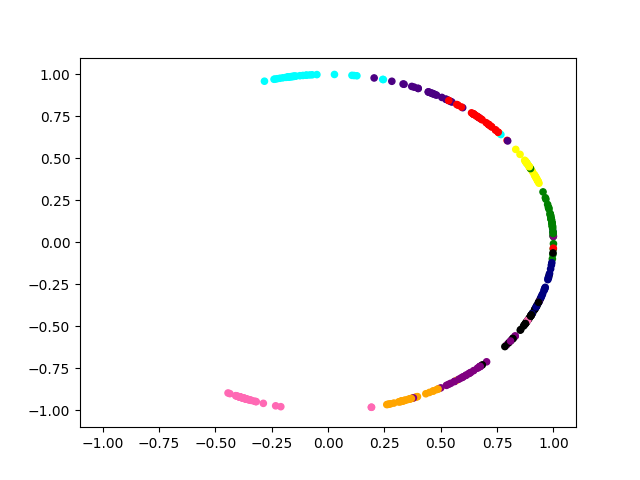} & \includegraphics[width=\svdAblation\textwidth,height=\svdAblation\textwidth]{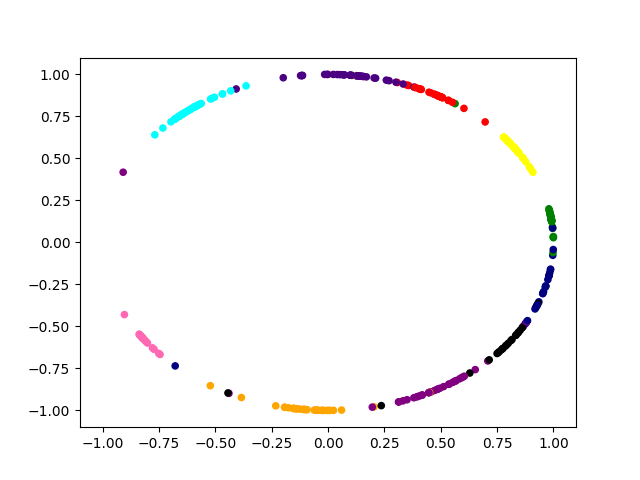} &
		\includegraphics[width=\svdAblation\textwidth,height=\svdAblation\textwidth]{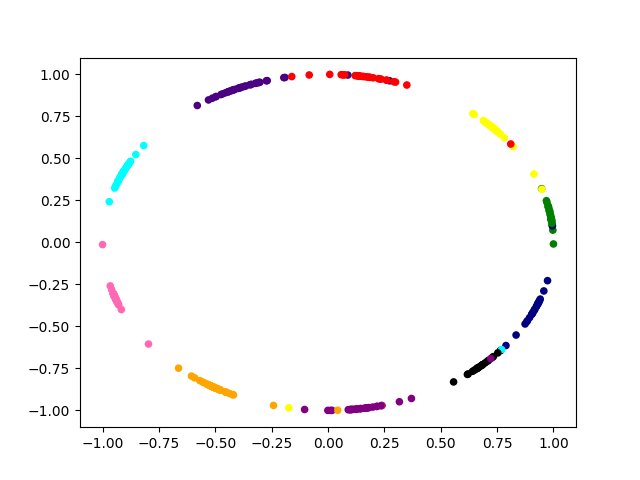} &
		\includegraphics[width=\svdAblation\textwidth,height=\svdAblation\textwidth]{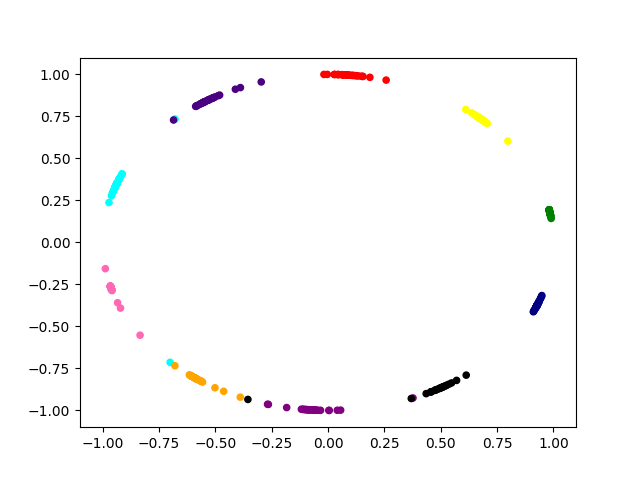} &
		\includegraphics[width=\svdAblation\textwidth,height=\svdAblation\textwidth]{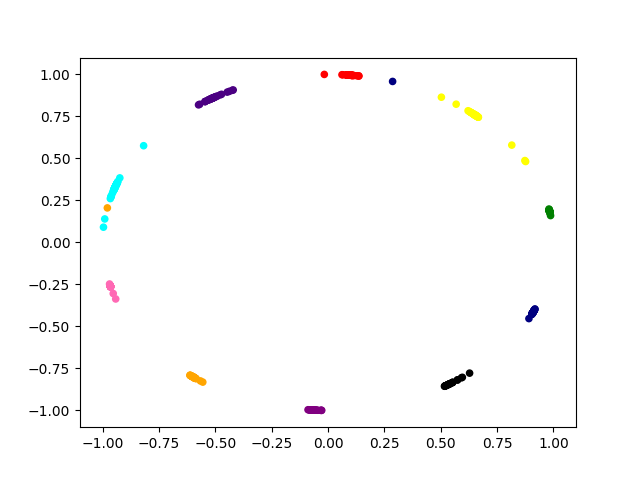} &
		\includegraphics[width=\svdAblation\textwidth,height=\svdAblation\textwidth]{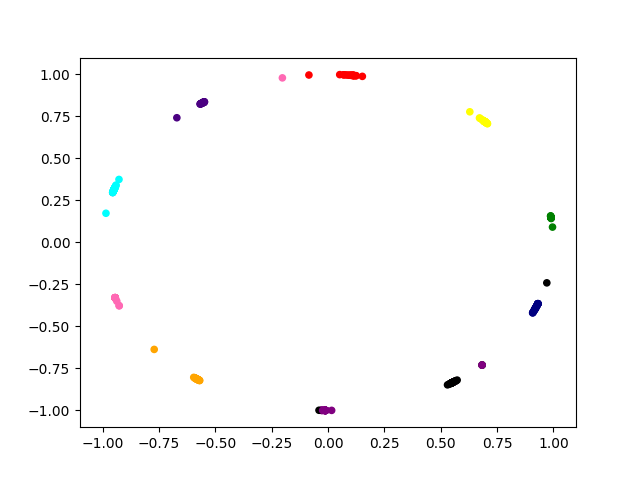} 
		\\
		
		\bottomrule
	\end{tabular}
	\caption{Qualitative feature embedding evaluation using the MNIST dataset projected onto the 2D unit circle. The first row shows the feature embedding learned using a vanilla contrastive loss and the second row applies the SVMax regularizer. A random subset of the test split is projected for visualization purpose. Different colors denote different classes. The regularized feature embedding spreads out uniformly and rapidly. The supplementary material shows the feature embedding evolves vividly up to 200 epochs. This high resolution figure is best seen on a screen.}
	\label{fig:svd_ablation}
\end{figure*}
\subsection{Generative Adversarial Networks}\label{sec:gans}

Model collapse is one of the main challenges of training generative adversarial networks (GANs)~\cite{metz2016unrolled,srivastava2017veegan,mao2019mode,salimans2016improved}.
To tackle this challenge,~\citet{metz2016unrolled} propose an unrolled-GAN to prevent the generator from overfitting to the discriminator. In an unrolled-GAN, the generator observes the discriminator for $l$ steps before updating the generator's parameters using the gradient from the final step. Alternatively, we leverage the simpler SVMax regularizer to avoid model collapse. We evaluate our regularizer using a simple GAN on a 2D mixture of 8 Gaussians arranged in a circle. This 2D baseline~\cite{metz2016unrolled,srivastava2017veegan,bang2018mggan} provides a simple qualitative evaluation and demonstrates SVMax's potential in unsupervised learning. We leverage this simple baseline because we assume $b\ge d$, which does not hold for images.


Figure~\ref{fig:qual_gan_one_colunm} shows the dynamics of the GAN generator through time. We use a public PyTorch implementation\footnote{https://github.com/andrewliao11/unrolled-gans} of~\cite{metz2016unrolled}. We made a single modification to the code to use a relatively large learning rate,~\ie, $lr=0.025$ for both the generator and discriminator. This single modification is a simple and fast way to induce model collapse. The mixture of Gaussians circle has a radius $r=2$,~\ie, the generated fake embedding is neither L2-normalized nor strictly bounded by a network layer. We kept the radius parameter unchanged to emphasize that neither L2-normalization nor strict-bounds are required. To mitigate the impact of lurking variables (\eg, random network initialization and mini-batch sampling), we fix the random generator's seed for all experiments. We apply SVMax to a vanilla and an unrolled GAN for five steps. We apply the unbounded SVMax regularizer (Eq.~\ref{eq:svd_avg_s}),~\ie,~$L_{\text{NN}} = L_{\text{GAN}} - \lambda s_\mu$, where $\lambda = 0.01$ and $s_\mu$ is mean singular value of the generator fake embedding.

%

GANs are typically used to generate high resolution images. This high-resolution output is the main limitation of the SVMax regularizer. The \textit{current} formulation assumes the batch size is bigger than the embedding dimension,~\ie, $b\ge d$. This constraint is trivial for the Gaussians mixture 2D dataset and retrieval networks with a compact embedding dimensionality (\eg, $d=\{128,256\}$). However, this constraint hinders high resolution image generators because the mini-batch size constraint becomes $b\ge W\times H \times C$, where $W$, $H$, and $C$ are the generated image's width, height, and number of channels, respectively. Nevertheless, this GAN experiment emphasizes the potential of the SVMax regularizer in unsupervised learning.

If the batch-size limitation is set aside, the following points are worth noting:  (\rom{1}) Image-synthesis GANs have bounded outputs [0, 255]; White images will not fool the discriminator. Thus, $s_\mu$ remains bounded but with different bounds from those presented in the approach section. (\rom{2}) \citet{alain2016understanding} ($\S$3.4) address practical concerns when working with high dimensional features. (\rom{3}) GANs have synthesized not only high quality images, but also feature embeddings~\cite{zhu2018generative}.


\subsection{Ablation Study}
In this section, we evaluate two hypotheses: (1) the same SVMax hyperparameter $\lambda$ supports different embedding dimensions and batch sizes -- the main objective of the mean singular value's bounds analysis, (2) the SVMax regularizer boosts retrieval performance because it learns a uniform feature embedding.


The mean singular value bound analysis makes tuning the hyperparameter $\lambda$ easier. This hyperparameter becomes only dependent on the ranking loss's range and independent of both the batch size and the embedding dimension. Figure~\ref{fig:quan_cub_mobile} presents a quantitative evaluation using the CUB-200 dataset. We explore various batch sizes $b=\{288,72\}$ and embedding dimensions $d=\{256,64\}$. We employ a MobileNetV2~\cite{sandler2018mobilenetv2} to fit the big batch $b=288$ on a 24GB GPU. The paper appendix contains a similar evaluation on the Stanford Online Products and CARS196 datasets.

To evaluate SVMax's impact on feature embeddings, we embed the MNIST dataset onto the 2D unit circle. In this experiment, we use a tiny CNN (one convolutional layer and one hidden layer). Figure~\ref{fig:svd_ablation} shows the feature embedding after training for $t$ epochs. With SVMax, the feature embeddings spread out more uniformly and rapidly than the vanilla contrastive loss.

\section{Conclusion}
We have proposed singular value maximization (SVMax) as a feature embedding regularizer. SVMax promotes a uniform embedding, mitigates model collapse, and enables large learning rates. Unlike other embedding regularizers, SVMax supports a large spectrum of ranking losses. Moreover, it is oblivious to data annotation and, as such, supports both supervised and unsupervised learning. Qualitative evaluation using a generative adversarial network demonstrates SVMax's potential in unsupervised learning. Quantitative retrieval evaluation highlight significant performance improvements due to the SVMax regularizer.


\noindent\textbf{Acknowledgments:} This work was partially funded by independent grants from Office of Naval Research (N000141612713) and Facebook AI. AH was supported by the NDSEG fellowship.

\bibliography{svmax}

\begin{thebibliography}{63}
\providecommand{\natexlab}[1]{#1}
\providecommand{\url}[1]{\texttt{#1}}
\expandafter\ifx\csname urlstyle\endcsname\relax
  \providecommand{\doi}[1]{doi: #1}\else
  \providecommand{\doi}{doi: \begingroup \urlstyle{rm}\Url}\fi

\bibitem[Alain \& Bengio(2016)Alain and Bengio]{alain2016understanding}
Alain, G. and Bengio, Y.
\newblock Understanding intermediate layers using linear classifier probes.
\newblock \emph{arXiv preprint arXiv:1610.01644}, 2016.

\bibitem[Bang \& Shim(2018)Bang and Shim]{bang2018mggan}
Bang, D. and Shim, H.
\newblock Mggan: Solving mode collapse using manifold guided training.
\newblock \emph{arXiv preprint arXiv:1804.04391}, 2018.

\bibitem[Caron et~al.(2020)Caron, Misra, Mairal, Goyal, Bojanowski, and
  Joulin]{caron2020unsupervised}
Caron, M., Misra, I., Mairal, J., Goyal, P., Bojanowski, P., and Joulin, A.
\newblock Unsupervised learning of visual features by contrasting cluster
  assignments.
\newblock \emph{arXiv preprint arXiv:2006.09882}, 2020.

\bibitem[Chen \& Deng(2019)Chen and Deng]{chen2019energy}
Chen, B. and Deng, W.
\newblock Energy confused adversarial metric learning for zero-shot image
  retrieval and clustering.
\newblock In \emph{AAAI}, 2019.

\bibitem[Chen et~al.(2020)Chen, Kornblith, Norouzi, and Hinton]{chen2020simple}
Chen, T., Kornblith, S., Norouzi, M., and Hinton, G.
\newblock A simple framework for contrastive learning of visual
  representations.
\newblock In \emph{ICML}, 2020.

\bibitem[Chen \& He(2020)Chen and He]{chen2020exploring}
Chen, X. and He, K.
\newblock Exploring simple siamese representation learning.
\newblock \emph{arXiv preprint arXiv:2011.10566}, 2020.

\bibitem[Deng et~al.(2009)Deng, Dong, Socher, Li, Li, and
  Fei-Fei]{deng2009imagenet}
Deng, J., Dong, W., Socher, R., Li, L.-J., Li, K., and Fei-Fei, L.
\newblock Imagenet: A large-scale hierarchical image database.
\newblock In \emph{CVPR}, 2009.

\bibitem[Doersch et~al.(2015)Doersch, Gupta, and
  Efros]{doersch2015unsupervised}
Doersch, C., Gupta, A., and Efros, A.~A.
\newblock Unsupervised visual representation learning by context prediction.
\newblock In \emph{ICCV}, 2015.

\bibitem[Donahue et~al.(2016)Donahue, Kr{\"a}henb{\"u}hl, and
  Darrell]{donahue2016adversarial}
Donahue, J., Kr{\"a}henb{\"u}hl, P., and Darrell, T.
\newblock Adversarial feature learning.
\newblock \emph{arXiv preprint arXiv:1605.09782}, 2016.

\bibitem[Friedland \& Lim(2016)Friedland and Lim]{friedland2016computational}
Friedland, S. and Lim, L.-H.
\newblock The computational complexity of duality.
\newblock \emph{SIAM Journal on Optimization}, 26\penalty0 (4):\penalty0
  2378--2393, 2016.

\bibitem[Goodfellow et~al.(2014)Goodfellow, Pouget-Abadie, Mirza, Xu,
  Warde-Farley, Ozair, Courville, and Bengio]{goodfellow2014generative}
Goodfellow, I., Pouget-Abadie, J., Mirza, M., Xu, B., Warde-Farley, D., Ozair,
  S., Courville, A., and Bengio, Y.
\newblock Generative adversarial nets.
\newblock In \emph{NIPS}, 2014.

\bibitem[Grill et~al.(2020)Grill, Strub, Altch{\'e}, Tallec, Richemond,
  Buchatskaya, Doersch, Pires, Guo, Azar, et~al.]{grill2020bootstrap}
Grill, J.-B., Strub, F., Altch{\'e}, F., Tallec, C., Richemond, P.~H.,
  Buchatskaya, E., Doersch, C., Pires, B.~A., Guo, Z.~D., Azar, M.~G., et~al.
\newblock Bootstrap your own latent: A new approach to self-supervised
  learning.
\newblock \emph{arXiv preprint arXiv:2006.07733}, 2020.

\bibitem[Guo \& Ye(2019)Guo and Ye]{guo2019regularization}
Guo, P. and Ye, Q.
\newblock On regularization for a convolutional kernel in neural networks.
\newblock \emph{arXiv preprint arXiv:1906.04866}, 2019.

\bibitem[Hadsell et~al.(2006)Hadsell, Chopra, and
  LeCun]{hadsell2006dimensionality}
Hadsell, R., Chopra, S., and LeCun, Y.
\newblock Dimensionality reduction by learning an invariant mapping.
\newblock In \emph{CVPR}, 2006.

\bibitem[He et~al.(2016)He, Zhang, Ren, and Sun]{he2016deep}
He, K., Zhang, X., Ren, S., and Sun, J.
\newblock Deep residual learning for image recognition.
\newblock In \emph{CVPR}, 2016.

\bibitem[He et~al.(2019)He, Fan, Wu, Xie, and Girshick]{he2019momentum}
He, K., Fan, H., Wu, Y., Xie, S., and Girshick, R.
\newblock Momentum contrast for unsupervised visual representation learning.
\newblock \emph{arXiv preprint arXiv:1911.05722}, 2019.

\bibitem[He et~al.(2018)He, Zhou, Zhou, Bai, and Bai]{he2018triplet}
He, X., Zhou, Y., Zhou, Z., Bai, S., and Bai, X.
\newblock Triplet-center loss for multi-view 3d object retrieval.
\newblock \emph{arXiv preprint arXiv:1803.06189}, 2018.

\bibitem[Hermans et~al.(2017)Hermans, Beyer, and Leibe]{hermans2017defense}
Hermans, A., Beyer, L., and Leibe, B.
\newblock In defense of the triplet loss for person re-identification.
\newblock \emph{arXiv preprint arXiv:1703.07737}, 2017.

\bibitem[Hinton et~al.(2015)Hinton, Vinyals, and Dean]{hinton2015distilling}
Hinton, G., Vinyals, O., and Dean, J.
\newblock Distilling the knowledge in a neural network.
\newblock \emph{arXiv preprint arXiv:1503.02531}, 2015.

\bibitem[Hoffer \& Ailon(2015)Hoffer and Ailon]{hoffer2015deep}
Hoffer, E. and Ailon, N.
\newblock Deep metric learning using triplet network.
\newblock In \emph{International Workshop on Similarity-Based Pattern
  Recognition}, 2015.

\bibitem[Hoffman et~al.(2019)Hoffman, Roberts, and Yaida]{hoffman2019robust}
Hoffman, J., Roberts, D.~A., and Yaida, S.
\newblock Robust learning with jacobian regularization.
\newblock \emph{arXiv preprint arXiv:1908.02729}, 2019.

\bibitem[Horn \& Johnson(1991)Horn and Johnson]{horn1991topics}
Horn, R.~A. and Johnson, C.~R.
\newblock Topics in matrix analysis cambridge university press.
\newblock \emph{Cambridge, UK}, 1991.

\bibitem[Hu(2015)]{hu2015relations}
Hu, S.
\newblock Relations of the nuclear norm of a tensor and its matrix flattenings.
\newblock \emph{Linear Algebra and its Applications}, 478:\penalty0 188--199,
  2015.

\bibitem[Ioffe \& Szegedy(2015)Ioffe and Szegedy]{ioffe2015batch}
Ioffe, S. and Szegedy, C.
\newblock Batch normalization: Accelerating deep network training by reducing
  internal covariate shift.
\newblock \emph{arXiv preprint arXiv:1502.03167}, 2015.

\bibitem[Ionescu et~al.(2015)Ionescu, Vantzos, and
  Sminchisescu]{ionescu2015training}
Ionescu, C., Vantzos, O., and Sminchisescu, C.
\newblock Training deep networks with structured layers by matrix
  backpropagation.
\newblock \emph{arXiv preprint arXiv:1509.07838}, 2015.

\bibitem[Kim et~al.(2020)Kim, Kim, Cho, and Kwak]{kim2020proxy}
Kim, S., Kim, D., Cho, M., and Kwak, S.
\newblock Proxy anchor loss for deep metric learning.
\newblock In \emph{CVPR}, 2020.

\bibitem[Kliegl et~al.(2017)Kliegl, Goyal, Zhao, Srinet, and
  Shoeybi]{kliegl2017trace}
Kliegl, M., Goyal, S., Zhao, K., Srinet, K., and Shoeybi, M.
\newblock Trace norm regularization and faster inference for embedded speech
  recognition rnns.
\newblock \emph{arXiv preprint arXiv:1710.09026}, 2017.

\bibitem[Kong et~al.(2018)Kong, Li, and Wang]{kong2018new}
Kong, X., Li, J., and Wang, X.
\newblock New estimations on the upper bounds for the nuclear norm of a tensor.
\newblock \emph{Journal of inequalities and applications}, 2018\penalty0
  (1):\penalty0 282, 2018.

\bibitem[Krause et~al.(2013)Krause, Stark, Deng, and Fei-Fei]{krause20133d}
Krause, J., Stark, M., Deng, J., and Fei-Fei, L.
\newblock 3d object representations for fine-grained categorization.
\newblock In \emph{Proceedings of the IEEE international conference on computer
  vision workshops}, 2013.

\bibitem[Kumar et~al.(2016)Kumar, Carneiro, Reid, et~al.]{kumar2016learning}
Kumar, B., Carneiro, G., Reid, I., et~al.
\newblock Learning local image descriptors with deep siamese and triplet
  convolutional networks by minimising global loss functions.
\newblock In \emph{Proceedings of the IEEE Conference on Computer Vision and
  Pattern Recognition}, pp.\  5385--5394, 2016.

\bibitem[Li et~al.(2016)Li, Kadav, Durdanovic, Samet, and Graf]{li2016pruning}
Li, H., Kadav, A., Durdanovic, I., Samet, H., and Graf, H.~P.
\newblock Pruning filters for efficient convnets.
\newblock \emph{arXiv preprint arXiv:1608.08710}, 2016.

\bibitem[Luo et~al.(2017)Luo, Wu, and Lin]{luo2017thinet}
Luo, J.-H., Wu, J., and Lin, W.
\newblock Thinet: A filter level pruning method for deep neural network
  compression.
\newblock In \emph{ICCV}, 2017.

\bibitem[Mao et~al.(2019)Mao, Lee, Tseng, Ma, and Yang]{mao2019mode}
Mao, Q., Lee, H.-Y., Tseng, H.-Y., Ma, S., and Yang, M.-H.
\newblock Mode seeking generative adversarial networks for diverse image
  synthesis.
\newblock In \emph{CVPR}, 2019.

\bibitem[Metz et~al.(2017)Metz, Poole, Pfau, and
  Sohl-Dickstein]{metz2016unrolled}
Metz, L., Poole, B., Pfau, D., and Sohl-Dickstein, J.
\newblock Unrolled generative adversarial networks.
\newblock In \emph{ICLR}, 2017.

\bibitem[Noroozi \& Favaro(2016)Noroozi and Favaro]{noroozi2016unsupervised}
Noroozi, M. and Favaro, P.
\newblock Unsupervised learning of visual representations by solving jigsaw
  puzzles.
\newblock In \emph{ECCV}, 2016.

\bibitem[Noroozi et~al.(2017)Noroozi, Pirsiavash, and
  Favaro]{noroozi2017representation}
Noroozi, M., Pirsiavash, H., and Favaro, P.
\newblock Representation learning by learning to count.
\newblock In \emph{ICCV}, 2017.

\bibitem[Oh~Song et~al.(2016)Oh~Song, Xiang, Jegelka, and Savarese]{oh2016deep}
Oh~Song, H., Xiang, Y., Jegelka, S., and Savarese, S.
\newblock Deep metric learning via lifted structured feature embedding.
\newblock In \emph{CVPR}, 2016.

\bibitem[Oord et~al.(2018)Oord, Li, and Vinyals]{oord2018representation}
Oord, A. v.~d., Li, Y., and Vinyals, O.
\newblock Representation learning with contrastive predictive coding.
\newblock \emph{arXiv preprint arXiv:1807.03748}, 2018.

\bibitem[Pathak et~al.(2016)Pathak, Krahenbuhl, Donahue, Darrell, and
  Efros]{pathak2016context}
Pathak, D., Krahenbuhl, P., Donahue, J., Darrell, T., and Efros, A.~A.
\newblock Context encoders: Feature learning by inpainting.
\newblock In \emph{CVPR}, 2016.

\bibitem[Rippel et~al.(2015)Rippel, Paluri, Dollar, and
  Bourdev]{rippel2015metric}
Rippel, O., Paluri, M., Dollar, P., and Bourdev, L.
\newblock Metric learning with adaptive density discrimination.
\newblock \emph{arXiv preprint arXiv:1511.05939}, 2015.

\bibitem[Sablayrolles et~al.(2018)Sablayrolles, Douze, Schmid, and
  J{\'e}gou]{sablayrolles2018spreading}
Sablayrolles, A., Douze, M., Schmid, C., and J{\'e}gou, H.
\newblock Spreading vectors for similarity search.
\newblock \emph{arXiv preprint arXiv:1806.03198}, 2018.

\bibitem[Salimans et~al.(2016)Salimans, Goodfellow, Zaremba, Cheung, Radford,
  and Chen]{salimans2016improved}
Salimans, T., Goodfellow, I., Zaremba, W., Cheung, V., Radford, A., and Chen,
  X.
\newblock Improved techniques for training gans.
\newblock In \emph{NIPS}, 2016.

\bibitem[Sandler et~al.(2018)Sandler, Howard, Zhu, Zhmoginov, and
  Chen]{sandler2018mobilenetv2}
Sandler, M., Howard, A., Zhu, M., Zhmoginov, A., and Chen, L.-C.
\newblock Mobilenetv2: Inverted residuals and linear bottlenecks.
\newblock In \emph{CVPR}, 2018.

\bibitem[Sanyal et~al.(2019)Sanyal, Kanade, and Torr]{amartya2019learning}
Sanyal, A., Kanade, V., and Torr, P. H.~S.
\newblock Learning low-rank representations.
\newblock \emph{arXiv preprint arXiv:1804.07090}, 2019.

\bibitem[Schroff et~al.(2015)Schroff, Kalenichenko, and
  Philbin]{schroff2015facenet}
Schroff, F., Kalenichenko, D., and Philbin, J.
\newblock Facenet: A unified embedding for face recognition and clustering.
\newblock In \emph{CVPR}, 2015.

\bibitem[Sedghi et~al.(2018)Sedghi, Gupta, and Long]{sedghi2018singular}
Sedghi, H., Gupta, V., and Long, P.~M.
\newblock The singular values of convolutional layers.
\newblock \emph{arXiv preprint arXiv:1805.10408}, 2018.

\bibitem[Sohn(2016)]{sohn2016improved}
Sohn, K.
\newblock Improved deep metric learning with multi-class n-pair loss objective.
\newblock In \emph{NIPS}, 2016.

\bibitem[Srivastava et~al.(2017)Srivastava, Valkov, Russell, Gutmann, and
  Sutton]{srivastava2017veegan}
Srivastava, A., Valkov, L., Russell, C., Gutmann, M.~U., and Sutton, C.
\newblock Veegan: Reducing mode collapse in gans using implicit variational
  learning.
\newblock In \emph{NSIP}, 2017.

\bibitem[Szegedy et~al.(2015)Szegedy, Liu, Jia, Sermanet, Reed, Anguelov,
  Erhan, Vanhoucke, and Rabinovich]{szegedy2015going}
Szegedy, C., Liu, W., Jia, Y., Sermanet, P., Reed, S., Anguelov, D., Erhan, D.,
  Vanhoucke, V., and Rabinovich, A.
\newblock Going deeper with convolutions.
\newblock In \emph{CVPR}, 2015.

\bibitem[Taha et~al.(2020)Taha, Chen, Misu, Shrivastava, and
  Davis]{taha2020boosting}
Taha, A., Chen, Y.-T., Misu, T., Shrivastava, A., and Davis, L.
\newblock Boosting standard classification architectures through a ranking
  regularizer.
\newblock In \emph{WACV}, 2020.

\bibitem[Tian et~al.(2019)Tian, Krishnan, and Isola]{tian2019contrastive}
Tian, Y., Krishnan, D., and Isola, P.
\newblock Contrastive multiview coding.
\newblock \emph{arXiv preprint arXiv:1906.05849}, 2019.

\bibitem[Turkmen \& Civciv(2007)Turkmen and Civciv]{turkmen2007some}
Turkmen, R. and Civciv, H.
\newblock Some bounds for the singular values of matrices.
\newblock \emph{Applied Mathematical Sciences}, 1\penalty0 (49):\penalty0
  2443--2449, 2007.

\bibitem[Wah et~al.(2011)Wah, Branson, Welinder, Perona, and
  Belongie]{wah2011caltech}
Wah, C., Branson, S., Welinder, P., Perona, P., and Belongie, S.
\newblock The caltech-ucsd birds-200-2011 dataset.
\newblock 2011.

\bibitem[Wang et~al.(2017)Wang, Zhou, Wen, Liu, and Lin]{wang2017deep}
Wang, J., Zhou, F., Wen, S., Liu, X., and Lin, Y.
\newblock Deep metric learning with angular loss.
\newblock In \emph{ICCV}, 2017.

\bibitem[Wen et~al.(2016)Wen, Zhang, Li, and Qiao]{wen2016discriminative}
Wen, Y., Zhang, K., Li, Z., and Qiao, Y.
\newblock A discriminative feature learning approach for deep face recognition.
\newblock In \emph{ECCV}, 2016.

\bibitem[Wu et~al.(2017)Wu, Manmatha, Smola, and Krahenbuhl]{wu2017sampling}
Wu, C.-Y., Manmatha, R., Smola, A.~J., and Krahenbuhl, P.
\newblock Sampling matters in deep embedding learning.
\newblock In \emph{ICCV}, 2017.

\bibitem[Xuan et~al.(2020)Xuan, Stylianou, and Pless]{xuan2020improved}
Xuan, H., Stylianou, A., and Pless, R.
\newblock Improved embeddings with easy positive triplet mining.
\newblock In \emph{WACV}, 2020.

\bibitem[Yu et~al.(2018)Yu, Li, Chen, Lai, Morariu, Han, Gao, Lin, and
  Davis]{yu2018nisp}
Yu, R., Li, A., Chen, C.-F., Lai, J.-H., Morariu, V.~I., Han, X., Gao, M., Lin,
  C.-Y., and Davis, L.~S.
\newblock Nisp: Pruning networks using neuron importance score propagation.
\newblock In \emph{CVPR}, 2018.

\bibitem[Zhang et~al.(2018)Zhang, Lei, and Dhillon]{zhang2018stabilizing}
Zhang, J., Lei, Q., and Dhillon, I.~S.
\newblock Stabilizing gradients for deep neural networks via efficient svd
  parameterization.
\newblock \emph{arXiv preprint arXiv:1803.09327}, 2018.

\bibitem[Zhang et~al.(2016)Zhang, Isola, and Efros]{zhang2016colorful}
Zhang, R., Isola, P., and Efros, A.~A.
\newblock Colorful image colorization.
\newblock In \emph{ECCV}, 2016.

\bibitem[Zhang et~al.(2017{\natexlab{a}})Zhang, Isola, and
  Efros]{zhang2017split}
Zhang, R., Isola, P., and Efros, A.~A.
\newblock Split-brain autoencoders: Unsupervised learning by cross-channel
  prediction.
\newblock In \emph{CVPR}, 2017{\natexlab{a}}.

\bibitem[Zhang et~al.(2017{\natexlab{b}})Zhang, Yu, Kumar, and
  Chang]{zhang2017learning}
Zhang, X., Yu, F.~X., Kumar, S., and Chang, S.-F.
\newblock Learning spread-out local feature descriptors.
\newblock In \emph{ICCV}, 2017{\natexlab{b}}.

\bibitem[Zhu et~al.(2018)Zhu, Elhoseiny, Liu, Peng, and
  Elgammal]{zhu2018generative}
Zhu, Y., Elhoseiny, M., Liu, B., Peng, X., and Elgammal, A.
\newblock A generative adversarial approach for zero-shot learning from noisy
  texts.
\newblock In \emph{CVPR}, 2018.

\end{thebibliography}
\bibliographystyle{icml2021}

\clearpage
The following appendix sections~\ref{sec:extend_approach} and~\ref{sec:supp_exp} extend their corresponding sections in the main paper. For instance, the extended-approach appendix~\ref{sec:extend_approach} extends the approach section in the main paper.
\appendix
\section{Appendix: Extended Approach}\label{sec:extend_approach}
This section presents the lower and upper bounds of N-pair and angular losses and analyzes a \textit{theoretical} corner case for SVMax.

\subsection{N-pair and Angular Losses}\label{sec:loss_analysis}
\textbf{Lower and Upper Bounds of Ranking Losses:} The N-pair and angular losses are bounded when their feature embeddings are L2-normalized. These bounds depend on the number of negative samples inside the training mini-batch $B$. Each mini-batch contains a single positive pair of samples per class, $\{a, p\}$, with all remaining samples, $n$, representing negative samples of that class. $a,p,n$ denote the anchor, positive and negative samples, respectively. This gives $|n| = b - 2$ as the number of negative samples w.r.t. a mini-batch of size $b$. Both losses also use the inner product operation to quantify similarity between feature embeddings,~\ie, $\in [-1,1]$, allowing us to find our bounds. Equation~\ref{eq:npair_loss} shows the N-pair loss (NL) formulation:
\begin{equation}\label{eq:npair_loss}
\text{NL} = -\log{\frac{exp(\lfloor a \rfloor \lfloor p \rfloor)}{exp(\lfloor a \rfloor \lfloor p \rfloor)+\sum_{n\in B}^{}exp(\lfloor a \rfloor \lfloor n \rfloor)}},
\end{equation}
where each $n$ is a negative sample, $a$ is the anchor, and $p$ is the positive sample inside the mini-batch $B$. $\lfloor \sbullet[0.75] \rfloor$ is the embedding function (encoder). Thus, the lower and upper bounds for the L2-normalized N-pair loss are the following:
\begin{equation}
[L,U]_\text{NL} = [\log{(e^2 + |n|)} - 2, \log{(e^2 |n| + 1)}].
\end{equation}
Equations~\ref{eq:svd_ang1} and~\ref{eq:svd_ang2} show the angular loss (AL) formulation:
\begin{align}
	\label{eq:svd_ang1} 
\text{AL} = \log{\left[ 1+\sum_{n\in B}^{}{exp(f_{a,p,n})}\right] },  \\ \nonumber
s.t.\quad f_{a,p,n} = 4\tan^2{\alpha}(\lfloor a \rfloor + \lfloor p \rfloor)^T \lfloor n \rfloor  \\ \label{eq:svd_ang2} - 2(1+\tan^2{\alpha}) {\lfloor a \rfloor}^T \lfloor p \rfloor,
\end{align}

where, again, each $n$ is a negative sample, $a$ is the anchor, $p$ is the positive sample inside the mini-batch $B$, and $\lfloor \sbullet[0.75] \rfloor$ is the embedding function (encoder). The parameter $\alpha$ is a hyperparameter chosen before training and is thus a fixed value. As such, the lower and upper bounds for L2-normalized $f_{a,p,n}$ are:
\begin{equation}
[L,U]_{f_{a,p,n}} =  \left[ -10\tan^2{\alpha} - 2, 6\tan^2{\alpha} - 2\right].
\end{equation}
We use $\alpha=45^{\circ}$ in all our experiments. This gives  $\tan{\alpha}=1, [L,U]_{f_{a,p,n}}\in[-12,4]$ and our angular loss bounds as:
\begin{equation}
[L,U]_{AL} =  \left[\log{(e^{-12}|n| + 1)} ,\log{( e^{4}|n| + 1) } \right].
\end{equation}

\subsection{SVMax Corner Case}\label{sec:svmax_corner_case}

\begin{figure}[t]
	\centering
	\includegraphics[width=0.5\linewidth]{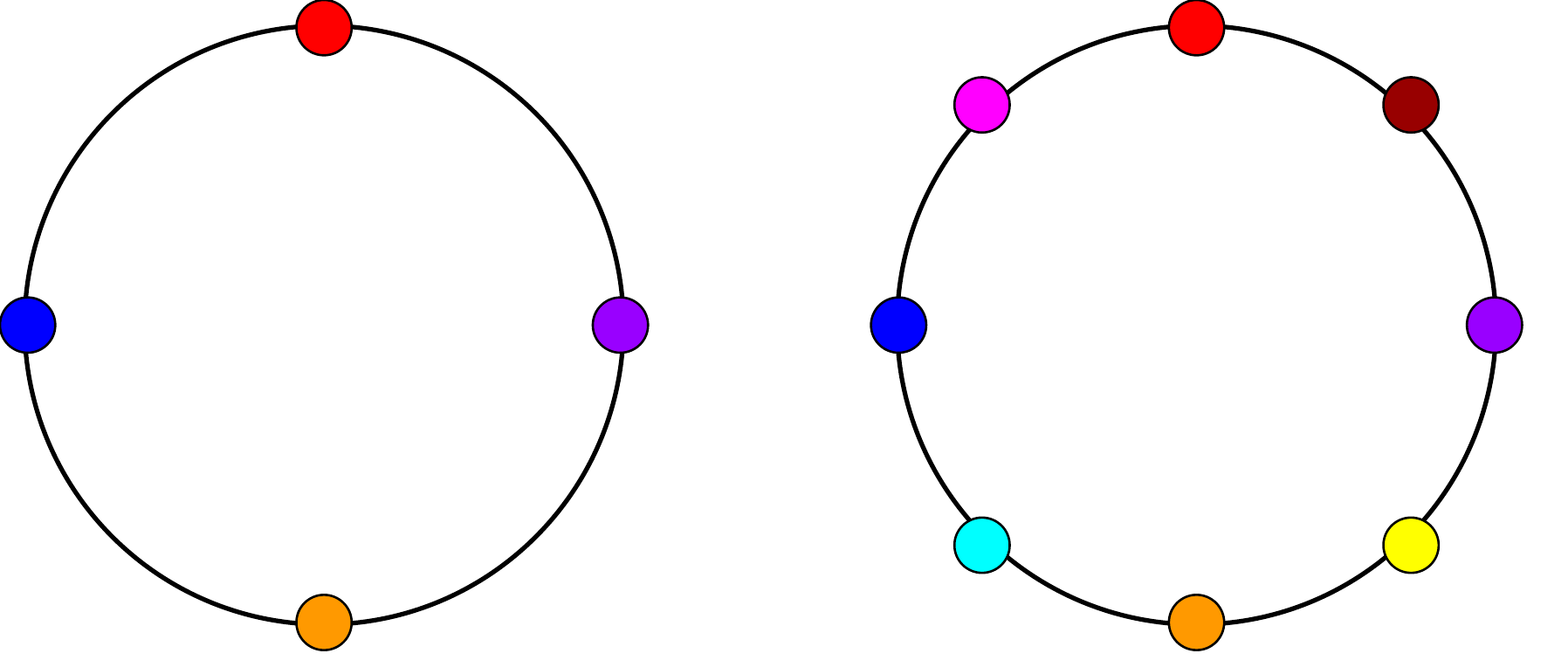}
	\caption{The feature embeddings of two independent mini-batches with $p=4$ (Left) and $p=8$ (Right) classes on the 2D unit circle. Different colors denote different classes. The mean singular value $s_\mu$ is maximum if (1) samples from the same class have zero standard deviation and (2) the $p$ sampled classes are distributed perfectly in the embedding space.}
	\label{fig:svmaxcornercase}
\end{figure}

Theoretically, the mean singular value $s_\mu$ can reach its upper bound for a mini-batch, even if the feature embedding is not perfectly uniform. During training, each mini-batch contains $p$ different classes and $l$ different samples per class,~\ie, the batch size $b=p\times l$. The sampled classes $p$ in a mini-batch is smaller than the number of total classes $C$ in the dataset,~\ie, $p<C$. If (1) the sampled $p$ classes are perfectly distributed in the feature embedding and (2) the $l$ samples for each class have zero standard deviation, the mean singular value equals the upper bound ($s_\mu=U$) even if the feature embedding for the whole dataset is not uniform. Figure~\ref{fig:svmaxcornercase} illustrates this scenario.

In practice, this will never happen during training because (1) the feature embedding dimension is large enough (\eg, $d=128$), (2) samples in the training mini-batches are randomly sampled, and (3) samples are independent across mini-batches. This corner case is omitted from the main paper because it undermines the paper’s flow and clarity.

\section{Appendix: Extended Experiments}\label{sec:supp_exp}
Section~\ref{sec:app_retrieval} provides further quantitative evaluations for SVMax when $b\ge d$. Then, Section~\ref{sec:svmax_small_minibatch} evaluates SVMax with small mini-batches,~\ie, $b<d$. Finally, Section~\ref{sec:rep_cnt} evaluates SVMax using self-supervised learning. The supplementary \textit{video} vividly shows how SVMax speeds convergence on the MNIST dataset.

\subsection{Retrieval Networks}\label{sec:app_retrieval}

\begin{figure}[t]
	\centering
	\scriptsize
	\begin{tikzpicture}
	\begin{groupplot}[group style = {group size = 2 by 2, horizontal sep = 20pt}, 
	height=3.7cm,
	symbolic x coords={0.0001,0.001,0.01},
	xtick=data,
	x label style={at={(axis description cs:0.5,-0.05)},anchor=north}
	]
	\nextgroupplot[title=Angular,ylabel=R@1,		y label style={at={(axis description cs:0.15,.5)}},
	legend style = { legend columns = -1, legend to name = grouplegend,}]
	\addplot[	color=blue,	dashed,	]
	coordinates {
		(0.0001,51.29979743) 
		(0.001,33.74409183) 
		(0.01,26.0972316) 
	}; \addlegendentry{\vanilla}
	\addplot[color=red,dashed,mark=*,]
	coordinates {
	(0.0001,51.33355841) 
	(0.001,34.04794058) 
	(0.01,28.13977043) 
}; \addlegendentry{\spread}

	\addplot[color=red,solid,]
	coordinates {
	(0.0001,50.99594868) 
	(0.001,35.3646185) 
	(0.01,29.18636057) 
}; \addlegendentry{SVMax}

		\nextgroupplot[title=N-pair,	legend style = { legend columns = -1, legend to name = grouplegend,}]
	\addplot[
	color=blue,	dashed,	]
	coordinates {
		(0.0001,41.86360567) 
		(0.001,33.74409183) 
		(0.01,23.70020257) 
	}; \addlegendentry{\vanilla}
	
	\addplot[	color=red,	dashed,	mark=*,]
	coordinates {
		(0.0001,44.05806887) 
		(0.001,35.60094531) 
		(0.01,27.46455098) 
	}; \addlegendentry{\spread}
	
	\addplot[	color=red,	solid,	]
	coordinates {
		(0.0001,44.19311276) 
		(0.001,47.09655638) 
		(0.01,36.03983795) 
	}; \addlegendentry{SVMax}	
	
			\nextgroupplot[title=Triplet,
				ylabel=R@1,y label style={at={(axis description cs:0.15,.5)}},
				legend style = { legend columns = -1, legend to name = grouplegend,}]
	\addplot[	color=blue,	dashed,	]
	coordinates {
		(0.0001,37.84604997) 
		(0.001,50.25320729) 
		(0.01,43.38284943) 
	}; \addlegendentry{\vanilla}
	
	\addplot[	color=red,	dashed,	mark=*,]
	coordinates {
		(0.0001,41.47535449) 
		(0.001,51.02970966) 
		(0.01,44.48008103) 
	}; \addlegendentry{\spread}
	
	\addplot[	color=red,	solid,]
	coordinates {
		(0.0001,39.97299122) 
		(0.001,50.59081702) 
		(0.01,44.56448346) 
	}; \addlegendentry{SVMax}	

	\nextgroupplot[title=Contrastive,	legend style = { legend columns = -1, legend to name = grouplegend,}]
	\addplot[	color=blue,	dashed,	]
	coordinates {
		(0.0001,35.58406482) 
		(0.001,41.64415935) 
		(0.01,30.06414585) 
	}; \addlegendentry{\vanilla}
	
	\addplot[	color=red,	dashed,	mark=*,]
	coordinates {
		(0.0001,43.07900068) 
		(0.001,37.15395003) 
		(0.01,28.1735314) 
	}; \addlegendentry{\spread}
	
	\addplot[	color=red,	solid,]
	coordinates {
		(0.0001,40.34436192) 
		(0.001,49.3923025) 
		(0.01,40.27683997) 
	}; \addlegendentry{SVMax}

	\end{groupplot}
	\node[below] at ($(group c1r2.south) +(1.75,-0.5)$) {\pgfplotslegendfromname{grouplegend}}; 
	\end{tikzpicture}
	
	\caption{Quantitative evaluation on CUB-200-2011 using ResNet50. The X axis denotes the learning rate $lr$ and the Y-axis denotes recall@1 performance. }
	\label{fig:quan_cub_resent}
\end{figure}
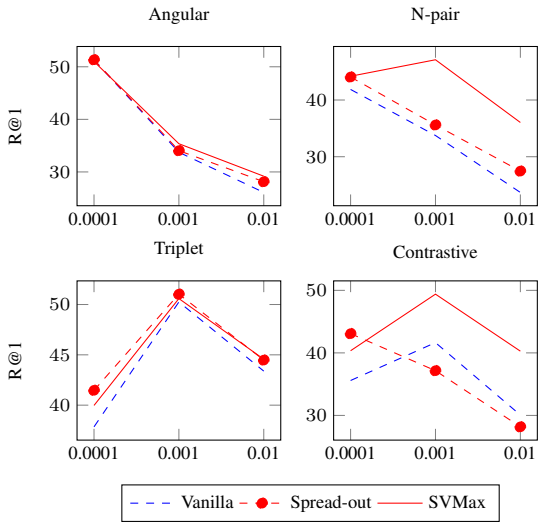
\begin{figure}[t]
	\centering
	\scriptsize
	\begin{tikzpicture}
	\begin{groupplot}[group style = {group size = 2 by 2, horizontal sep = 20pt}, 
	height=3.7cm,
	symbolic x coords={0.0001,0.001,0.01},
	xtick=data,
	x label style={at={(axis description cs:0.5,-0.05)},anchor=north}
	]
	\nextgroupplot[title=Angular,ylabel=R@1,		y label style={at={(axis description cs:0.15,.5)}},
	legend style = { legend columns = -1, legend to name = grouplegend,}]
	\addplot[	color=blue,	dashed,	]
	coordinates {
		(0.0001,57.6681835) 
		(0.001,49.60029517) 
		(0.01,35.48149059) 
	}; \addlegendentry{\vanilla}
	\addplot[color=red,dashed,mark=*,]
	coordinates {
	(0.0001,57.63128766) 
	(0.001,49.74787849) 
	(0.01,35.73976141) 
}; \addlegendentry{\spread}

	\addplot[color=red,solid,]
	coordinates {
	(0.0001,56.79498217) 
	(0.001,51.56807281) 
	(0.01,39.65071947) 
}; \addlegendentry{SVMax}

		\nextgroupplot[title=N-Pair,	legend style = { legend columns = -1, legend to name = grouplegend,}]
	\addplot[
	color=blue,	dashed,	]
	coordinates {
		(0.0001,34.35001845) 
		(0.001,36.03492805) 
		(0.01,32.99717132) 
	}; \addlegendentry{\vanilla}
	
	\addplot[	color=red,	dashed,mark=*,	]
	coordinates {
		(0.0001,39.47853893) 
		(0.001,45.38187185) 
		(0.01,35.92424056) 
	}; \addlegendentry{\spread}
	
	\addplot[	color=red,	solid,	]
	coordinates {
		(0.0001,40.47472636) 
		(0.001,64.12495388) 
		(0.01,53.49895462) 
	}; \addlegendentry{SVMax}	
	
			\nextgroupplot[title=Triplet,
				ylabel=R@1,y label style={at={(axis description cs:0.15,.5)}},
				legend style = { legend columns = -1, legend to name = grouplegend,}]
	\addplot[	color=blue,	dashed,	]
	coordinates {
		(0.0001,29.76263682) 
		(0.001,37.4123724) 
		(0.01,61.86200959) 
	}; \addlegendentry{\vanilla}
	
	\addplot[	color=red,	dashed,	mark=*,]
	coordinates {
		(0.0001,35.4937892) 
		(0.001,62.26786373) 
		(0.01,62.20637068) 
	}; \addlegendentry{\spread}
	
	\addplot[	color=red,	solid,]
	coordinates {
		(0.0001,31.06628951) 
		(0.001,46.94379535) 
		(0.01,65.50239823) 
	}; \addlegendentry{SVMax}	

	\nextgroupplot[title=Contrastive,	legend style = { legend columns = -1, legend to name = grouplegend,}]
	\addplot[	color=blue,	dashed,	]
	coordinates {
		(0.0001,30.30377567) 
		(0.001,37.91661542) 
		(0.01,35.56758086) 
	}; \addlegendentry{\vanilla}
	
	\addplot[	color=red,	dashed,	mark=*,]
	coordinates {
		(0.0001,36.99421965) 
		(0.001,43.57397614) 
		(0.01,35.56758086) 
	}; \addlegendentry{\spread}
	
	\addplot[	color=red,	solid,]
	coordinates {
		(0.0001,34.57139343) 
		(0.001,60.58295413) 
		(0.01,59.24240561) 
	}; \addlegendentry{SVMax}

	\end{groupplot}
	\node[below] at ($(group c1r2.south) +(1.75,-0.5)$) {\pgfplotslegendfromname{grouplegend}}; 
	\end{tikzpicture}
	
	\caption{Quantitative evaluation on Stanford CARS196 using ResNet50.}
	\label{fig:quan_cars_resent}
\end{figure}
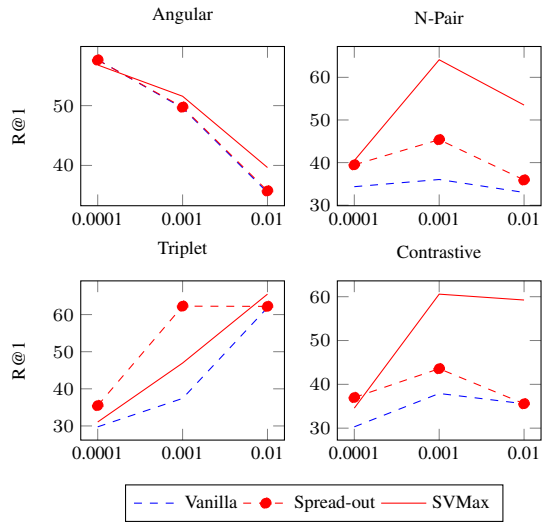

\begin{table*}[t]
	\centering
	\scriptsize
	\caption{Quantitative evaluation using N-pair loss without L2-normalization on three datasets and GoogLeNet with batch size $b=144$, embedding dimension $d=128$ and multiple learning rates $lr=\{0.01,0.001,0.0001\}$. $\triangle_{R@1}$ column indicates the R@1 improvement margin relative to the vanilla ranking loss. SVMax leverages the upper and lower bounds while $\text{SVMax}''$ does not.}
	\setlength{\tabcolsep}{3pt}
	\begin{tabular}{@{} l cccc c cccc c cccc   @{}}
		\toprule
		& \multicolumn{4}{c}{$lr=0.01$} && \multicolumn{4}{c}{$lr=0.001$} && \multicolumn{4}{c}{$lr=0.0001$}\\
		\cmidrule{2-5} \cmidrule{7-10} \cmidrule{12-15}
		Method & NMI & R@1 & R@8 & $\triangle_{R@1}$ && NMI & R@1 & R@8 & $\triangle_{R@1}$ && NMI & R@1 & R@8 & $\triangle_{R@1}$\\
		\midrule
		
		& \multicolumn{14}{c}{N-pair on CUB-200-2011}\\
		\cmidrule{2-15}
		
		Vanilla &       0.560      &  \recall{0.4608372721}   & \recall{0.7908507765}   & - &
		&       0.553      & \recall{0.4424375422}    & \recall{0.7812288994}  &  - &
		&         0.555     &   \recall{0.4532410533}   &   \recall{0.7871370695}  &  -      \\
		
		Spread-out &      \multicolumn{3}{c}{degenerate (Trn loss $\rightarrow$ nan)} & - &
		&  0.560  &  \recall{0.4579675895}    & \recall{0.786630655}         &    \margin{0.4579675895-0.4424375422} &
		& 0.537   &   \recall{0.4267386901}   &  \recall{0.7716070223}    & \margin{0.4267386901-0.4532410533}   \\
		
		$\text{SVMax}''$ (Ours) &  \bf 0.563       &   \recall{0.4579675895}   &\recall{0.7972653612}    & \margin{0.4579675895-0.4608372721} &
		&  0.553 &     \recall{0.4490209318}    & \recall{0.7857866307}     & \margin{0.4490209318-0.4424375422}  &
		&      0.555       &     \recall{0.4496961512}   &   \recall{0.7879810939}    &   \margin{0.4496961512-0.4532410533}   \\

		SVMax (Ours) &  0.561       &   \bf  \recall{0.4626941256}   &\bf  \recall{0.7986158001}    & \margin{0.4626941256-0.4608372721} &
		&  \bf  0.563 &   \bf   \recall{0.4640445645}    & \bf  \recall{0.7969277515}     & \margin{0.4640445645-0.4424375422}  &
		&      \bf  0.563       &   \bf   \recall{0.4569547603}   & \bf    \recall{0.79270763}    &   \margin{0.4569547603-0.4532410533}   \\
		
		\midrule
		& \multicolumn{14}{c}{N-pair on Stanford CARS196} \\
		\cmidrule{2-15}
		
		Vanilla &       0.606      &  \recall{0.6549009962}   & \recall{0.9032099373}   & - &
		&        0.5983      &  \recall{0.6492436355}    & \recall{0.9000122986}  &  - &
		&         0.480    &   \recall{0.441274136}   &  \recall{0.7976878613}  &  -      \\
		
		Spread-out &      \multicolumn{3}{c}{degenerate (Trn loss $\rightarrow$ nan)} & - &
		&  0.109  &  \recall{0.02582708154}    & \recall{0.1406961013}         &    \margin{0.02582708154-0.6492436355} &
		& 0.393   &   \recall{0.315336367}   &  \recall{0.6785143279}    & \margin{0.315336367-0.441274136}   \\
		
		$\text{SVMax}''$  (Ours) &  0.606       &   \recall{0.6545320379}   & \recall{0.9087443119}    & \margin{0.6545320379-0.6549009962} &
		& 0.592 &     \recall{0.6492436355}    & \recall{0.9007502152}     & \margin{0.6492436355-0.6492436355}  &
		&      0.481       &      \recall{0.442503997}   &    \recall{0.7992866806}    &   \margin{0.442503997-0.441274136}   \\
		
		SVMax (Ours) & \bf  0.611       &   \bf  \recall{0.6972082155}   & \bf  \recall{0.9215348666}    & \margin{0.6972082155-0.6549009962} &
		&  \bf  0.601 &   \bf   \recall{0.6731029394}    & \bf  \recall{0.9132947977}     & \margin{0.6731029394-0.6492436355}  &
		&    \bf   0.492       &    \bf   \recall{0.4872709384}   &   \bf  \recall{0.8317550117}    &   \margin{0.4872709384-0.441274136}   \\
		
		\midrule
		& \multicolumn{14}{c}{N-pair on Stanford Online Products} \\
		\cmidrule{2-15}
		
		Vanilla &    \bf    0.897      &   \recall{0.7453968464}   &  \recall{0.8792106046}   & - &
		&       \bf  0.884      &  \recall{0.6934481505}    & \recall{0.8445836501}  &  - &
		&       0.863    &  \recall{0.596757132}   &   \recall{0.7641565568}  &  -      \\
		
		Spread-out &      \multicolumn{3}{c}{degenerate (Trn loss $\rightarrow$ nan)} & - &
		&  0.877  &  \recall{0.6423424019}    & \recall{0.8033949291}         &    \margin{0.6423424019-0.6934481505} &
		& 0.858   &   \recall{0.5649565304}   &  \recall{0.7315130078}    & \margin{0.5649565304-0.596757132}   \\
		
		$\text{SVMax}''$ (Ours) & \bf  0.897       &  \bf  \recall{0.7481736141}   & \bf  \recall{0.8829956035}    & \margin{0.7481736141-0.7453968464} &
		&  0.883 &     \recall{0.6929853559}    &  \recall{0.844881161}     & \margin{0.6929853559-0.6934481505}  &
		&   \bf  0.864       &      \recall{0.596889359}   &    \recall{0.764338369}    &   \margin{0.596889359-0.596757132}   \\
		
		SVMax (Ours) &  0.895       &   \recall{0.7444051436}   &\recall{0.8790618492}    & \margin{0.7444051436-0.7453968464} &
		&  0.882 &   \bf   \recall{0.6951505735}    & \bf  \recall{0.846236488}     & \margin{0.6951505735-0.6934481505}  &
		&    \bf   0.864       &    \bf  \recall{0.5993355592}   &  \bf  \recall{0.767346534}    &   \margin{0.5993355592-0.596757132}   \\
		\bottomrule
	\end{tabular}
	\label{tbl:npair_cub}
\end{table*}

\topic{Evaluation Metrics} For quantitative evaluation, we leverage the \textbf{Recall@K} metric and \textbf{Normalized Mutual Info} (NMI) on the test split. The NMI score evaluates the quality of cluster alignments. $\text{NMI}=\frac { I(\Omega ,C) }{ \sqrt { H(\Omega )H(C) }  } ,$ where $\Omega =\{\omega_1,..,\omega_n\}$, is the ground-truth clustering, while $C=\{c_1,...c_n\}$ is a clustering assignment for the learned embedding. $I(\sbullet[0.5],\sbullet[0.5])$ and $H(\sbullet[0.5])$ denote mutual information and entropy, respectively. We use K-means to compute $C$.

\textbf{Quantitative Evaluation:}  In the main paper, SVMax is evaluated quantitatively using GoogLeNet. Figures~\ref{fig:quan_cub_resent} and~\ref{fig:quan_cars_resent} present quantitative evaluation using ResNet50 on CUB-200-2011 and Stanford CARS196, respectively. 


Our evaluation hyperparameters (\eg, learning rate and batch size) do not favor a particular ranking loss.  The ideal hyperparameters depend on the ranking loss and other factors such as the dataset size, batch size, and network architecture. The hyperparameters in the main paper are inconsistent with the N-pair loss because this loss assumes an un-normalized embedding. Table~\ref{tbl:npair_cub} presents an SVMax evaluation using un-normalized embedding on three datasets. We leverage both the unbound SVMax ($\text{SVMax}''$) in equation~\ref{eq:svd_avg_s_supp} and the bounded SVMax (SVMax) in equation~\ref{eq:svd_u_l_supp}. Surprisingly, the bounded SVMax, which assumes an L2-normalized embedding, achieves competitive performance on the un-normalized embedding. It is important to note that while N-pair loss assumes an un-normalized embedding, N-pair loss \textit{regularizes} the L2-norm of the embedding vectors to be small. Table~\ref{tbl:npair_cub} shows that the spread-out regularizer degenerates severely, while SVMax remains resilient. The spread-out regularizer requires an L2-normalized embedding while SVMax does not. In Table~\ref{tbl:npair_cub}, the unbound SVMax uses $\lambda=0.01$ while the bounded SVMax uses $\lambda=1$.

\begin{figure*}[t]
	\centering
	\scriptsize
	\begin{tikzpicture}
	\begin{groupplot}[group style = {group size = 4 by 2, horizontal sep = 20pt}, 
	height=3.7cm,
	symbolic x coords={0.0001,0.001,0.01},
	xtick=data,
	x label style={at={(axis description cs:0.5,-0.05)},anchor=north}
	]
	\nextgroupplot[title=\texttt{Angular \textit{d=256}},ylabel=R@1,		y label style={at={(axis description cs:0.15,.5)}},
	legend style = { legend columns = -1, legend to name = grouplegend,}]
	\addplot[	color=blue,	dashed,	]
	coordinates {
		(0.0001,65.86393838) 
		(0.001,69.76794156) 
		(0.01,48.41492843) 
	}; \addlegendentry{\vanilla}
	\addplot[color=red,dashed,mark=*,]
	coordinates {
	(0.0001,65.87881392) 
	(0.001,69.88364021) 
	(0.01,47.6463588) 
}; \addlegendentry{\spread}

	\addplot[color=red,solid,]
	coordinates {
	(0.0001,66.81432019) 
	(0.001,70.97451324) 
	(0.01,55.56179961) 
}; \addlegendentry{SVMax}

		\nextgroupplot[title=\texttt{N-pair \textit{d=256}},	legend style = { legend columns = -1, legend to name = grouplegend,}]
	\addplot[
	color=blue,	dashed,	]
	coordinates {
		(0.0001,43.06469208) 
		(0.001,27.68999372) 
		(0.01,11.78638723) 
	}; \addlegendentry{\vanilla}
	
	\addplot[	color=red,	dashed,mark=*,	]
	coordinates {
		(0.0001,48.857889) 
		(0.001,39.01028065) 
		(0.01,16.17136624) 
	}; \addlegendentry{\spread}
	
	\addplot[	color=red,	solid,	]
	coordinates {
		(0.0001,58.34352583) 
		(0.001,65.48047998) 
		(0.01,52.100757) 
	}; \addlegendentry{SVMax}	
	
			\nextgroupplot[title=\texttt{Triplet \textit{d=256}},	legend style = { legend columns = -1, legend to name = grouplegend,}]
	\addplot[	color=blue,	dashed,	]
	coordinates {
		(0.0001,60.60295527) 
		(0.001,71.13318568) 
		(0.01,67.77461902) 
	}; \addlegendentry{\vanilla}
	
	\addplot[	color=red,	dashed,	mark=*,]
	coordinates {
		(0.0001,61.0310403) 
		(0.001,71.06541932) 
		(0.01,68.24071932) 
	}; \addlegendentry{\spread}
	
	\addplot[	color=red,	solid,]
	coordinates {
		(0.0001,60.53023041) 
		(0.001,70.81749364) 
		(0.01,67.52669333) 
	}; \addlegendentry{SVMax}	

	\nextgroupplot[title=\texttt{Contrastive \textit{d=256}},	legend style = { legend columns = -1, legend to name = grouplegend,}]
	\addplot[	color=blue,	dashed,	]
	coordinates {
		(0.0001,49.59174903) 
		(0.001,36.07649334) 
		(0.01,15.10693861) 
	}; \addlegendentry{\vanilla}
	
	\addplot[	color=red,	dashed,	mark=*,]
	coordinates {
		(0.0001,47.06125417) 
		(0.001,35.75418994) 
		(0.01,19.76463588) 
	}; \addlegendentry{\spread}
	
	\addplot[	color=red,	solid,]
	coordinates {
		(0.0001,57.13364847) 
		(0.001,65.66229216) 
		(0.01,59.65092063) 
	}; \addlegendentry{SVMax}

		\nextgroupplot[title=\texttt{Angular \textit{d=64}},ylabel=R@1,		y label style={at={(axis description cs:0.15,.5)}},
	legend style = { legend columns = -1, legend to name = grouplegend,}]
	\addplot[	color=blue,	dashed,	]
	coordinates {
	(0.0001,60.54510595) 
	(0.001,66.79944465) 
	(0.01,63.5417011) 
}; \addlegendentry{\vanilla}
\addplot[color=red,dashed,mark=*,]
coordinates {
	(0.0001,60.55998149) 
	(0.001,66.78787478) 
	(0.01,63.46897623) 
}; \addlegendentry{\spread}

\addplot[color=red,solid,]
coordinates {
	(0.0001,60.65749893) 
	(0.001,66.34987273) 
	(0.01,62.53016429) 
}; \addlegendentry{SVMax}

	\nextgroupplot[title=\texttt{N-pair \textit{d=64}},	legend style = { legend columns = -1, legend to name = grouplegend,}]
	\addplot[
	color=blue,	dashed,	]
	coordinates {
		(0.0001,34.43687812) 
		(0.001,26.0636012) 
		(0.01,17.72172821) 
	}; \addlegendentry{\vanilla}
	
	\addplot[	color=red,	dashed,	mark=*,]
	coordinates {
		(0.0001,42.72255463) 
		(0.001,40.05322138) 
		(0.01,29.41225083) 
	}; \addlegendentry{\spread}
	
	\addplot[	color=red,	solid,	]
	coordinates {
		(0.0001,50.41320948) 
		(0.001,57.90552378) 
		(0.01,51.97514132) 
	}; \addlegendentry{SVMax}	
	
	\nextgroupplot[title=\texttt{Triplet \textit{d=64}},	legend style = { legend columns = -1, legend to name = grouplegend,}]
	\addplot[	color=blue,	dashed,	]
	coordinates {
		(0.0001,50.59667449) 
		(0.001,64.58794751) 
		(0.01,69.73323196) 
	}; \addlegendentry{\vanilla}
	
	\addplot[	color=red,	dashed,	mark=*,]
	coordinates {
		(0.0001,52.21480282) 
		(0.001,64.91520941) 
		(0.01,66.78787478) 
	}; \addlegendentry{\spread}
	
	\addplot[	color=red,	solid,]
	coordinates {
		(0.0001,51.97183564) 
		(0.001,64.64910251) 
		(0.01,70.27205712) 
	}; \addlegendentry{SVMax}	
	
	\nextgroupplot[title=\texttt{Contrastive \textit{d=64}},	legend style = { legend columns = -1, legend to name = grouplegend,}]
	\addplot[	color=blue,	dashed,	]
	coordinates {
		(0.0001,37.55082477) 
		(0.001,29.410598) 
		(0.01,20.45056362) 
	}; \addlegendentry{\vanilla}
	
	\addplot[	color=red,	dashed,mark=*,	]
	coordinates {
		(0.0001,39.34745959) 
		(0.001,33.90962282) 
		(0.01,29.49654557) 
	}; \addlegendentry{\spread}
	
	\addplot[	color=red,	solid,]
	coordinates {
		(0.0001,47.23645499) 
		(0.001,57.59479025) 
		(0.01,56.94522495) 
	}; \addlegendentry{SVMax}

	\end{groupplot}
	
	\node[below] at ($(group c2r2.south) +(1.8,-0.5)$) {\pgfplotslegendfromname{grouplegend}}; 
	\end{tikzpicture}
	
	\caption{Quantitative evaluation on Stanford Online Products using various embedding dimensions $d=\{256,64\}$ to demonstrate the stability of our hyperparameter.}
	\label{fig:quan_stanford_mobile}
\end{figure*}
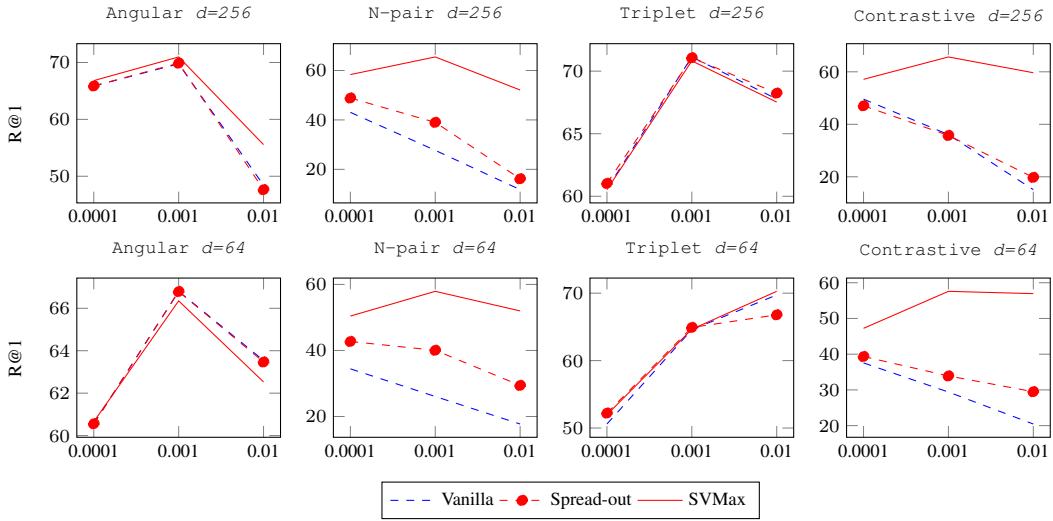
\begin{figure*}[t]
	\centering
	\scriptsize
	\begin{tikzpicture}
	\begin{groupplot}[group style = {group size = 3 by 2, horizontal sep = 20pt}, 
	height=3.7cm,
	symbolic x coords={0.0001,0.001,0.01},
	xtick=data,
	x label style={at={(axis description cs:0.5,-0.05)},anchor=north}
	]
	\nextgroupplot[title=\texttt{Triplet \textit{b=288,d=256}},ylabel=R@1,		y label style={at={(axis description cs:0.15,.5)}},
	legend style = { legend columns = -1, legend to name = grouplegend,}]
	\addplot[	color=blue,	dashed,	]
	coordinates {
		(0.0001,32.6159144) 
		(0.001,39.54003198) 
		(0.01,59.64825975) 
	}; \addlegendentry{\vanilla}
	\addplot[color=red,dashed,mark=*,]
	coordinates {
	(0.0001,33.47681712) 
	(0.001,50.00614931) 
	(0.01,61.65293322) 
}; \addlegendentry{\spread}

	\addplot[color=red,solid,]
	coordinates {
	(0.0001,32.67740745) 
	(0.001,44.37338581) 
	(0.01,62.9442873) 
}; \addlegendentry{SVMax}

		\nextgroupplot[title=\texttt{Triplet \textit{b=288,d=64}},	legend style = { legend columns = -1, legend to name = grouplegend,}]
	\addplot[
	color=blue,	dashed,	]
	coordinates {
		(0.0001,15.38556143) 
		(0.001,19.17353339) 
		(0.01,25.23674825) 
	}; \addlegendentry{\vanilla}
	
	\addplot[	color=red,	dashed,	mark=*,]
	coordinates {
		(0.0001,22.00221375) 
		(0.001,52.52736441) 
		(0.01,58.65207232) 
	}; \addlegendentry{\spread}
	
	\addplot[	color=red,	solid,	]
	coordinates {
		(0.0001,17.05817243) 
		(0.001,29.08621326) 
		(0.01,58.59057926) 
	}; \addlegendentry{SVMax}

		\nextgroupplot[title=\texttt{Triplet \textit{b=72,d=64}},	legend style = { legend columns = -1, legend to name = grouplegend,}]
	\addplot[
	color=blue,	dashed,	]
	coordinates {
		(0.0001,17.29184602) 
		(0.001,24.32665109) 
		(0.01,52.17070471) 
	}; \addlegendentry{\vanilla}
	
	\addplot[	color=red,	dashed,	mark=*,]
	coordinates {
		(0.0001,24.49883163) 
		(0.001,57.59439183) 
		(0.01,62.5876276) 
	}; \addlegendentry{\spread}
	
	\addplot[	color=red,	solid,	]
	coordinates {
		(0.0001,19.07514451) 
		(0.001,39.23256672) 
		(0.01,66.38789817) 
	}; \addlegendentry{SVMax}

			\nextgroupplot[title=\texttt{Contrastive \textit{b=288,d=256}},	legend style = { legend columns = -1, legend to name = grouplegend,}]
	\addplot[	color=blue,	dashed,	]
	coordinates {
		(0.0001,31.5336367) 
		(0.001,38.38396261) 
		(0.01,33.34153241) 
	}; \addlegendentry{\vanilla}
	
	\addplot[	color=red,	dashed,	mark=*,]
	coordinates {
		(0.0001,36.85893494) 
		(0.001,43.0451359) 
		(0.01,38.371664) 
	}; \addlegendentry{\spread}
	
	\addplot[	color=red,	solid,]
	coordinates {
		(0.0001,33.31693519) 
		(0.001,53.88021154) 
		(0.01,44.17660804) 
	}; \addlegendentry{SVMax}	

	\nextgroupplot[title=\texttt{Contrastive \textit{b=288,d=64}},	legend style = { legend columns = -1, legend to name = grouplegend,}]
	\addplot[	color=blue,	dashed,	]
	coordinates {
		(0.0001,18.47251261) 
		(0.001,28.39749108) 
		(0.01,29.60275489) 
	}; \addlegendentry{\vanilla}
	
	\addplot[	color=red,	dashed,	mark=*,]
	coordinates {
		(0.0001,28.43438691) 
		(0.001,37.85512237) 
		(0.01,37.59685156) 
	}; \addlegendentry{\spread}
	
	\addplot[	color=red,	solid,]
	coordinates {
		(0.0001,23.15828311) 
		(0.001,58.35690567) 
		(0.01,51.23601033) 
	}; \addlegendentry{SVMax}	
	
		\nextgroupplot[title=\texttt{Contrastive \textit{b=72,d=64}},	legend style = { legend columns = -1, legend to name = grouplegend,}]
	\addplot[	color=blue,	dashed,	]
	coordinates {
		(0.0001,18.65699176) 
		(0.001,28.12692166) 
		(0.01,31.69351863) 
	}; \addlegendentry{\vanilla}
	
	\addplot[	color=red,	dashed,	mark=*,]
	coordinates {
		(0.0001,28.64346329) 
		(0.001,38.74062231) 
		(0.01,35.38310171) 
	}; \addlegendentry{\spread}
	
	\addplot[	color=red,	solid,]
	coordinates {
		(0.0001,22.21129012) 
		(0.001,54.33526012) 
		(0.01,59.20550978) 
	}; \addlegendentry{SVMax}	
	
	\end{groupplot}
	\node[below] at ($(group c2r2.south) +(0.0,-0.5)$) {\pgfplotslegendfromname{grouplegend}}; 
	\end{tikzpicture}
	
	\caption{Quantitative evaluation on Stanford CARS196 using MobileNet, various embedding dimensions $d=\{256,64\}$, and batch sizes $b=\{288,72\}$ to demonstrate the stability of our hyperparameter. $\lambda=1 \text{ and } 0.1$ for contrastive and triplet loss, respectively.}
	\label{fig:quan_cars_mobile}
\end{figure*}
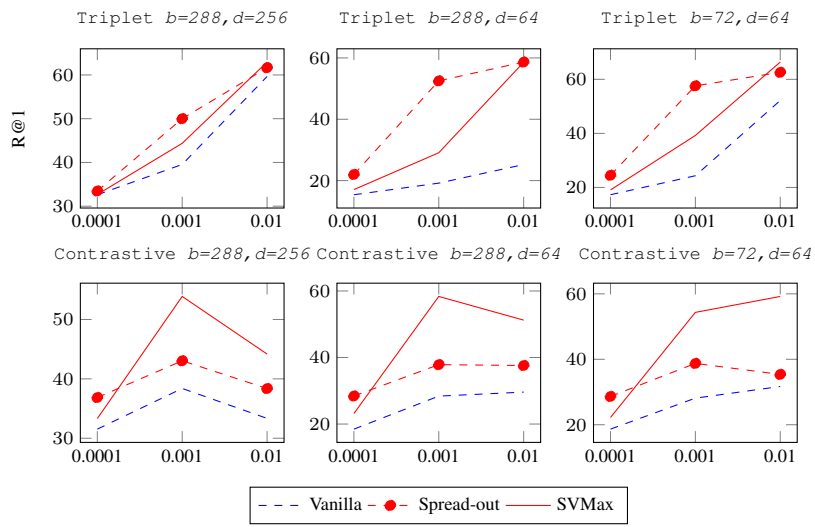




\begin{equation}\label{eq:svd_avg_s_supp}
L_{\text{NN}} = L_r - \lambda \frac{1}{d}\sum_{i=1}^{d}{s_i} = L_r - \lambda s_\mu,
\end{equation}
\begin{equation}\label{eq:svd_u_l_supp}
L_{\text{NN}} = L_r + \lambda \exp(\frac{U-s_\mu}{U-L} ).
\end{equation}

Figures~\ref{fig:quan_stanford_mobile} and~\ref{fig:quan_cars_mobile} present a quantitative evaluation with various embedding dimensions. We use batch sizes $b=\{288,72\}$ and embedding dimensions $d=\{256,64\}$. In this experiment, we employ a MobileNetV2~\cite{sandler2018mobilenetv2} to fit our neural network on a 24GB GPU. The N-pair and angular loss evaluations are dropped in Figure~\ref{fig:quan_cars_mobile} because these losses assume a single pair of anchor-positive per class. The CARS196, with 98 training classes, is too small to provide $144=\frac{288}{2}$ anchor-positive pairs.

In the main paper, we discuss two factors that contribute to model collapse in retrieval networks: learning rate and dataset intra-class variations. However, additional factors can also contribute. For instance, the likelihood of model collapse decreases as the mini-batch size $b$ increases. In the early training stages, a large learning rate will induce a noisy gradient. A large training mini-batch mitigates this noisy gradient and learns a better feature embedding.  The next section evaluates SVMax with small training mini-batches.






 \clearpage
\subsection{SVMax with Small Batches}\label{sec:svmax_small_minibatch}

\begin{figure}[t]
	\centering
	\scriptsize
	\begin{tikzpicture}
	\begin{groupplot}[group style = {group size = 2 by 2, horizontal sep = 20pt}, 
	height=3.7cm,
	symbolic x coords={0.0001,0.001,0.01},
	xtick=data,
	x label style={at={(axis description cs:0.5,-0.05)},anchor=north}
	]
	\nextgroupplot[title=Angular,ylabel=R@1,		y label style={at={(axis description cs:0.15,.5)}},
	legend style = { legend columns = -1, legend to name = grouplegend,}]
	\addplot[	color=blue,	dashed,	]
	coordinates {
		(0.0001,41.32343011) 
		(0.001,40.34436192) 
		(0.01,30.19918974) 
	}; \addlegendentry{\vanilla}
	\addplot[color=red,dashed,mark=*,]
	coordinates {
	(0.0001,41.72856178) 
	(0.001,40.95205942) 
	(0.01,29.86158001) 
}; \addlegendentry{\spread}

	\addplot[color=red,solid,]
	coordinates {
	(0.0001,41.28966914) 
	(0.001,43.39972991) 
	(0.01,37.15395003) 
}; \addlegendentry{SVMax}

		\nextgroupplot[title=N-pair,	legend style = { legend columns = -1, legend to name = grouplegend,}]
	\addplot[
	color=blue,	dashed,	]
	coordinates {
		(0.0001,31.60027009) 
		(0.001,27.66711681) 
		(0.01,19.17623228) 
	}; \addlegendentry{\vanilla}
	
	\addplot[	color=red,	dashed,	mark=*,]
	coordinates {
		(0.0001,33.32207968) 
		(0.001,33.23767725) 
		(0.01,23.68332208) 
	}; \addlegendentry{\spread}
	
	\addplot[	color=red,	solid,	]
	coordinates {
		(0.0001,33.67656989) 
		(0.001,42.09993248) 
		(0.01,35.8541526) 
	}; \addlegendentry{SVMax}	
	
			\nextgroupplot[title=Triplet,
				ylabel=R@1,		y label style={at={(axis description cs:0.15,.5)}},
				legend style = { legend columns = -1, legend to name = grouplegend,}]
	\addplot[	color=blue,	dashed,	]
	coordinates {
		(0.0001,29.15259959) 
		(0.001,39.92234976) 
		(0.01,44.83457124) 
	}; \addlegendentry{\vanilla}
	
	\addplot[	color=red,	dashed,	mark=*,]
	coordinates {
		(0.0001,30.35111411) 
		(0.001,44.04118839) 
		(0.01,44.05806887) 
	}; \addlegendentry{\spread}
	
	\addplot[	color=red,	solid,]
	coordinates {
		(0.0001,30.01350439) 
		(0.001,44.31127616) 
		(0.01,45.02025658) 
	}; \addlegendentry{SVMax}	

	\nextgroupplot[title=Contrastive,	legend style = { legend columns = -1, legend to name = grouplegend,}]
	\addplot[	color=blue,	dashed,	]
	coordinates {
		(0.0001,24.5104659) 
		(0.001,28.4098582) 
		(0.01,26.41796084) 
	}; \addlegendentry{\vanilla}
	
	\addplot[	color=red,	dashed,	mark=*,]
	coordinates {
		(0.0001,31.63403106) 
		(0.001,32.49493585) 
		(0.01,27.83592167) 
	}; \addlegendentry{\spread}
	
	\addplot[	color=red,	solid,]
	coordinates {
		(0.0001,29.81093856) 
		(0.001,42.26873734) 
		(0.01,42.28561783) 
	}; \addlegendentry{SVMax}

	\end{groupplot}
	\node[below] at ($(group c1r2.south) +(1.75,-0.5)$) {\pgfplotslegendfromname{grouplegend}}; 
	\end{tikzpicture}
	
	\caption{Quantitative evaluation on CUB-200-2011 using GoogLeNet with $b=72$ and $d=128$,~\ie, $b<d$.}
	\label{fig:quan_cub_inc_b}
\end{figure}
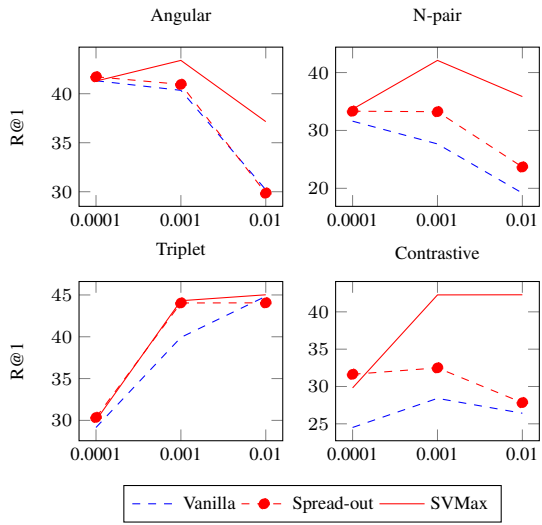
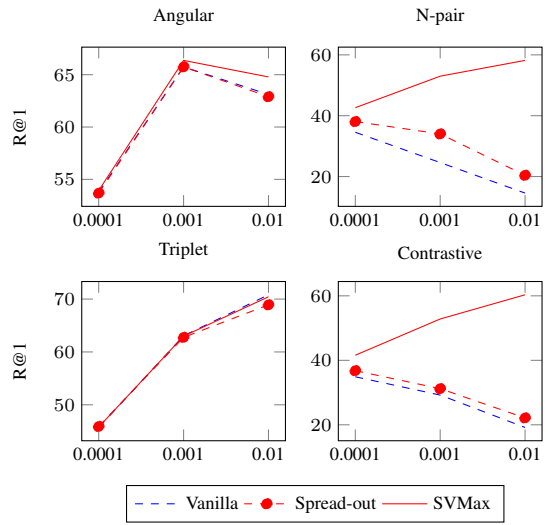
\begin{figure}[t]
	\centering
	\scriptsize
	\begin{tikzpicture}
	\begin{groupplot}[group style = {group size = 2 by 2, horizontal sep = 20pt}, 
	height=3.7cm,
	symbolic x coords={0.0001,0.001,0.01},
	xtick=data,
	x label style={at={(axis description cs:0.5,-0.05)},anchor=north}
	]
	\nextgroupplot[title=Angular,ylabel=R@1,		y label style={at={(axis description cs:0.15,.5)}},
	legend style = { legend columns = -1, legend to name = grouplegend,}]
	\addplot[	color=blue,	dashed,	]
	coordinates {
		(0.0001,53.77177614) 
		(0.001,65.74493405) 
		(0.01,63.1103104) 
	}; \addlegendentry{\vanilla}
	\addplot[color=red,dashed,mark=*,]
	coordinates {
	(0.0001,53.68748141) 
	(0.001,65.74658689) 
	(0.01,62.89874715) 
}; \addlegendentry{\spread}

	\addplot[color=red,solid,]
	coordinates {
	(0.0001,53.93705993) 
	(0.001,66.36805395) 
	(0.01,64.78794089) 
}; \addlegendentry{SVMax}

		\nextgroupplot[title=N-pair,	legend style = { legend columns = -1, legend to name = grouplegend,}]
	\addplot[
	color=blue,	dashed,	]
	coordinates {
		(0.0001,34.55918813) 
		(0.001,24.59588113) 
		(0.01,14.56480777) 
	}; \addlegendentry{\vanilla}
	
	\addplot[	color=red,	dashed,	mark=*,]
	coordinates {
		(0.0001,38.07642723) 
		(0.001,34.01540445) 
		(0.01,20.41420118) 
	}; \addlegendentry{\spread}
	
	\addplot[	color=red,	solid,	]
	coordinates {
		(0.0001,42.63495422) 
		(0.001,52.99824799) 
		(0.01,58.18485339) 
	}; \addlegendentry{SVMax}	
	
			\nextgroupplot[title=Triplet,
				ylabel=R@1,		y label style={at={(axis description cs:0.15,.5)}},
				legend style = { legend columns = -1, legend to name = grouplegend,}]
	\addplot[	color=blue,	dashed,	]
	coordinates {
		(0.0001,45.71253843) 
		(0.001,63.05411391) 
		(0.01,70.81253512) 
	}; \addlegendentry{\vanilla}
	
	\addplot[	color=red,	dashed,	mark=*,]
	coordinates {
		(0.0001,45.90426763) 
		(0.001,62.75660309) 
		(0.01,68.91177151) 
	}; \addlegendentry{\spread}
	
	\addplot[	color=red,	solid,]
	coordinates {
		(0.0001,45.79518032) 
		(0.001,62.99461175) 
		(0.01,70.43403524) 
	}; \addlegendentry{SVMax}	

	\nextgroupplot[title=Contrastive,	legend style = { legend columns = -1, legend to name = grouplegend,}]
	\addplot[	color=blue,	dashed,	]
	coordinates {
		(0.0001,34.89636706) 
		(0.001,29.22052164) 
		(0.01,19.17457274) 
	}; \addlegendentry{\vanilla}
	
	\addplot[	color=red,	dashed,	mark=*,]
	coordinates {
		(0.0001,36.793825) 
		(0.001,31.21053849) 
		(0.01,22.13811114) 
	}; \addlegendentry{\spread}
	
	\addplot[	color=red,	solid,]
	coordinates {
		(0.0001,41.58044362) 
		(0.001,52.8197415) 
		(0.01,60.3252785) 
	}; \addlegendentry{SVMax}

	\end{groupplot}
	\node[below] at ($(group c1r2.south) +(1.75,-0.5)$) {\pgfplotslegendfromname{grouplegend}}; 
	\end{tikzpicture}
	
	\caption{Quantitative evaluation on Stanford Online Products using GoogLeNet with $b=72$ and $d=128$,~\ie, $b<d$.}
	\label{fig:quan_stan_inc_b}
\end{figure}

 \begin{figure}[t]
	\centering
\begin{tikzpicture}
\begin{axis}[
ylabel=Mean Singular Value $(s_\mu)$,
xlabel=Batch Size,
enlargelimits=0.15,
legend style={at={(1.4,0.8)},
	anchor=east,legend columns=1},
ybar,
ymax=7,
width=2.5in,
xtick=data,
xmin=50,
xmax=166,
nodes near coords,
]
\addplot 
coordinates {(72,1.901554108) (144,1.899610281)};

\addplot 
coordinates {(72,5.637135983) (144,5.808442593) };

\addplot[red,sharp plot,update limits=false] 
coordinates {(0,6.803032412) (200,6.803032412)} 
node[above] at (axis cs:108,6.803032412) {Max $s_\mu$};

\legend{Vanila,SVMax}
\end{axis}
\end{tikzpicture}
	\caption{The mean singular values $s_\mu$ for networks trained with an embedding dimension $d=128$. The X and Y-axes denote the mini-batch size $b$ and the $s_\mu$ of  the feature embedding of CUB-200's test split.  The feature embedding is learned using a contrastive loss with and without SVMax. The horizontal red line denotes the upper bound on $s_\mu$. With SVMax,  $s_\mu$ decreases marginally with $b=72$ compared to $b=144$. Thus, the SVMax still promotes a uniform feature embedding even when $b<d$.}
\label{fig:max_mean_sv}
\end{figure}
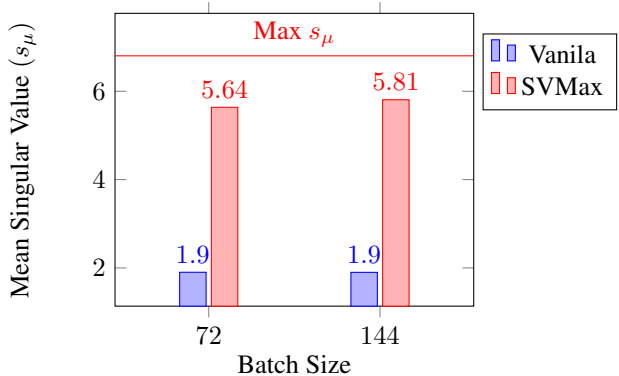

In the approach section, we assumed $b \ge d$ to deliver a rigorous mathematical foundation for SVMax. In this section, we present empirical evidence to support SVMax with small mini-batches. When $b<d$, there will be at most $b$ singular values, instead of $d$. The lower and upper bounds of SVMax, per mini-batch, become $[L,U] = [\frac{\sqrt{b}}{b},\sqrt{\frac{b\times d}{max(b,d)}}\frac{\sqrt{b}}{b}]$. It is possible that SVMax will utilize only $b$ dimensions of the feature embedding space. We argue against this possibility using a toy example. Consider the following two mini-batches ($m_1,m_2$) $\in R^{3\times d}$ 
\begin{align}
m_{1}&=\begin{bmatrix} m_{11} \\ m_{12} \\ m_{13}   \end{bmatrix}=\begin{bmatrix} 1 & 0 & 0 & 0 & .. & 0 \\ 0 & 1 & 0 & 0 & .. & 0 \\ 0 & 0 & 1 & 0 & .. & 0 \end{bmatrix} ,\\ 
m_{2}&=\begin{bmatrix} m_{21} \\ m_{22} \\ m_{23}   \end{bmatrix}=\begin{bmatrix} 1 & 0 & 0 & 0 & .. & 0 \\ 0 & 1 & 0 & 0 & .. & 0 \\ 0 & 0 & 1 & 0 & .. & 0 \end{bmatrix}, 
\end{align}
 where the mini-batch size $b=3$. Each individual mini-batch utilizes only the first three dimensions,~\ie, $rank(m_1)=rank(m_2)=3$. While all other dimensions $[4,d]$ contain zeros, the maximum mean singular value is feasible with only the first three dimensions. However, due to the random sampling procedure, a future mini-batch $m_3$ will contain elements from both $m_1$ and $m_2$. For instance, $m_3=[m_{11}\text{ }m_{21}\text{ }m_{22}]^T$ will have a $rank(m_3)=2$. For the mini-batch $m_3$, the mean singular value is not maximum. To maximize $s_\mu$, one feasible solution is to keep utilizing only the first three dimensions. However, this solution is like tossing a coin $N$ times and expecting $N$ heads. It is a feasible solution but unlikely.
 
 
 

 Figure~\ref{fig:quan_cub_inc_b} presents a quantitative evaluation using CUB-200 on GoogLeNet with $b=72$ and $d=128$. Similarly, Figure~\ref{fig:quan_stan_inc_b} presents a quantitative evaluation using Stanford Online Products. SVMax consistently outperforms the vanilla and spread-out baselines even when $b<d$. 

Finally, Figure~\ref{fig:max_mean_sv} depicts the mean singular value on the test split of CUB-200. We train our network using (1) contrastive loss with and without SVMax, and (2) different mini-batch sizes $b=\{72,144\}$. We fix the embedding dimension $d=128$ to study the batch size's impact,~\ie, $b<d$ versus $b\ge d$. The test split of CUB-200 has $5924$ test images. Thus, the upper bound of the mean singular value $U=\sqrt{\frac{b\times d}{max(b,d)}}\frac{\sqrt{b}}{d}=6.80$, where $b=5924$ and $d=128$ for the \textit{whole} test split. After training our network, the actual mean singular value $s_\mu = 5.64$ with batch size $b=72$, and  $s_\mu = 5.81$ with $b=144$. These mean singular values significantly outperform their vanilla contrastive loss counterparts ($s_\mu=1.9$). Compared to $b=144$, $s_\mu$ is smaller when using the mini-batch size $b=72$. At a mini-batch level, SVMax spreads the feature embedding across $d=128$ dimensions when $b=144$, while SVMax spreads the feature embedding across $d=72$ dimensions when $b=72$. Yet, the comparable $s_\mu$ (5.64 versus 5.81) indicates that SVMax supports $b<d$.

  \subsection{Self-Supervised Learning}~\label{sec:rep_cnt}
 
 Another form of model collapse is a loss function with a trivial solution. This form manifests in the two terms of contrastive loss~\autoref{eq:contrastive} as follows 
 \begin{equation}\label{eq:contrastive}
	\text{CL}_{(x,y)\in P} = { { { \delta_{x,y}D_{x,y} } }+{ (1 - \delta_{x,y})\left[ { m-D_{x,y} } \right]  }_{ + } },
 \end{equation} 
where $D_{x_1,x_2}=D(N(x_1),N(x_2))$; $N(\sbullet[0.75]) $ and $D(\sbullet[0.75],\sbullet[0.75])$ are the network's output-embedding and Euclidean distance, respectively. While the first term pulls similar points together, the second term pushes different points apart. Without the second term, contrastive loss suffers a model collapse,~\ie the trivial solution $N(x)=0, \; \forall x$. In supervised metric learning, similar and different classes are labeled. Thus, it is trivial to leverage contrastive loss with both terms. However, it is non-trivial to leverage contrastive loss in un/self-supervised learning.


To avoid this trivial solution, different methods have been proposed for un/self-supervised learning. For example, SimCLR~\cite{chen2020simple} repels random images -- assuming they belong to different classes. Other approaches like SwAV~\cite{caron2020unsupervised} leverages online clustering, while BYOL~\cite{grill2020bootstrap} leverages a momentum encoder. Recently, SimSiam~\cite{chen2020exploring} utilize a stop-gradient operation to avoid model collapse. All these methods deliver SOTA results, but they make assumptions about the problem formulation. For example, both SimCLR and SwAV require a large batch (e.g., 4096) to work well. In the following experiment, we show that SVMax avoids model collapse without making assumptions about the problem formulation.

\begin{figure}[t]
	\centering
	\scriptsize
	\includegraphics[width=0.35\textwidth]{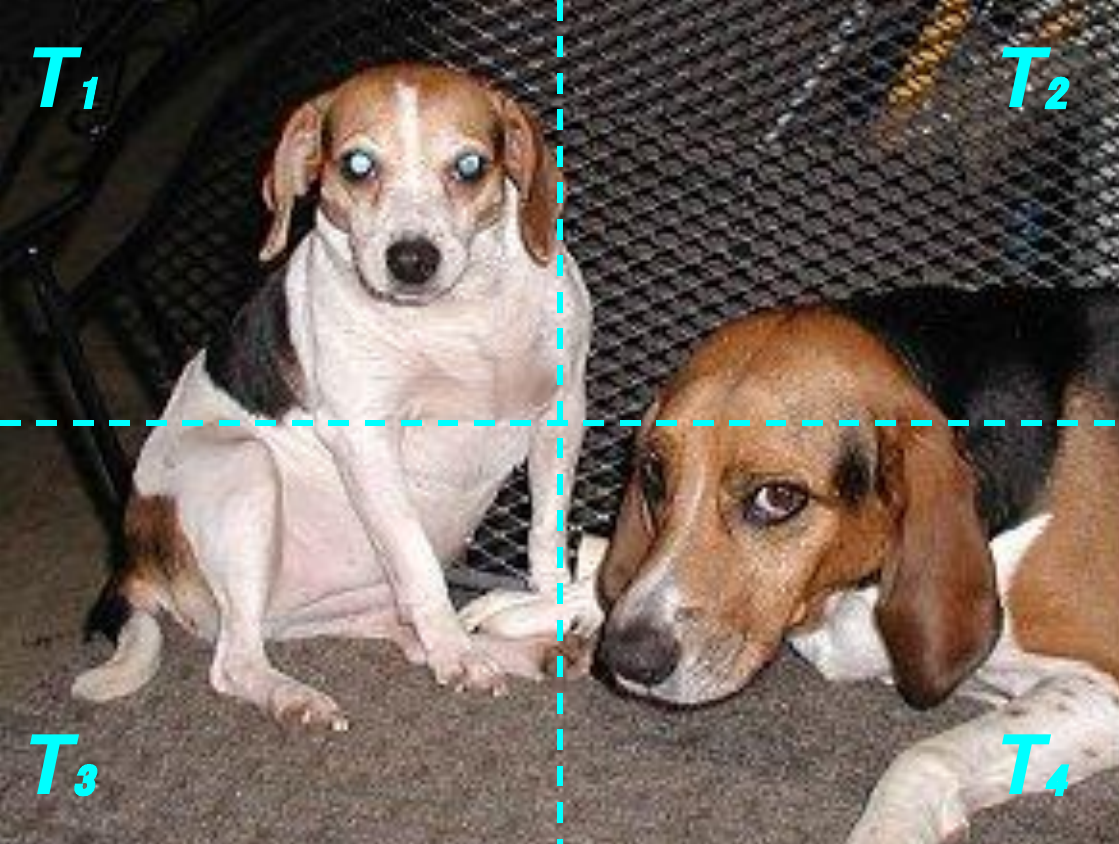}
	\vspace{0.05in}
	
	\begin{tabular}{@{} l cccc @{}}
		\toprule
		Tile & Nose & Eye & Ear & Paws \\
		\midrule
		$T_1$ & 1 & 2 & 2 & 0 \\
		$T_2$ & 0 & 0 & 0 & 0 \\
		$T_3$ & 0 & 0 & 0 & 3 \\
		$T_4$ & 1 & 2 & 2 & 1 \\
		\midrule
		$I$ (Total) & 2 & 4 & 4 & 4 \\
		\bottomrule
	\end{tabular}

	\caption{An image $I$ is split into four tiles $T_i$, where $i\in \{1,2,3,4\}$. To learn an image representation, Rep-Cnt trains a network $N$ such that counts of visual primitives in $I$ equals the total visual primitives in  $T_i$ as shown in the table -- $N(I) = \sum_{i=1}^4{N(T_i)}$.}\label{fig:rep_cnt}
\end{figure} 

To evaluate SVMax in self-supervised learning, we first introduce representation counting (Rep-Cnt)~\cite{noroozi2017representation}. Rep-Cnt is a simple self-supervised approach that counts visual primitives. Given an input image $I$, Rep-Cnt splits $I$ into four tiles $T_i$, where $i\in \{1,2,3,4\}$ as shown in~\autoref{fig:rep_cnt}. Rep-Cnt trains a neural network $N$ to count visual primitives in $I$ and $T_i$. The network is trained to maintain the equivariance relation. The equivariance relation requires that the count of visual primitives in $I$ equals the total visual primitives from the four tiles,~\ie $N(I) = \sum_{i=1}^4{N(T_i)}$. Unfortunately, this simple self-supervised signal has a trivial solution,~\ie $N(x) =0, \; \forall x$. To avoid the trivial solution, Rep-Cnt is formulated as follows

\begin{equation}\label{eq:rep_cnt}
	L =  \mathbb{D}_{I,T_i} +\underbrace{\left[ m-\mathbb{D}_{\hat{I},T_i} \right]_+}_{Problem-specific}
\end{equation}
where $\mathbb{D}_{I,T_i} = D\left(N(I) , \sum_{i=1}^4{N(T_i)}\right)$. The second term pushes the representation of a random image $\hat{I}$ from $\left(\sum_{i=1}^4{N(T_i)}\right)$,~\ie the representation of $I$. This loss formulation explains how a problem-specific loss term is always required to avoid model collapse. Similarly, a different self-supervised pretext requires a different problem-specific formulation. Instead, we propose SVMax, a generic prior, to promote a uniform feature embedding.

 \begin{table}[t]
	\centering
	\scriptsize
	\setlength{\tabcolsep}{3pt}
	\caption{Quantitative SVMax evaluation using self-supervised learning. We evaluate the pretrained network $N$ through ImageNet classification with a linear classifier on top of frozen convolutional layers. For every layer, the convolutional features are spatially resized until there are fewer than 10K dimensions left. A fully connected layer followed by softmax is trained on a 1000-way object classification task. \textbf{*} denotes our implementation of the baseline.}
	\begin{tabular}{@{}lccccc@{}}
		\toprule
		Method     & conv1 & conv2 & conv3 & conv4 & conv5 \\
		\midrule
		Supervised & 19.3  & 36.3  & 44.2  & 48.3  & 50.5  \\
		Random     & 11.6  & 17.1  & 16.9  & 16.3  & 14.1  \\
		\midrule
		Context~\cite{doersch2015unsupervised}  & 16.2 & 23.3 & 30.2 & 31.7 & 29.6 \\
		Jigsaw~\cite{noroozi2016unsupervised} & 18.2 & 28.8 &   34.0 & 33.9 & 27.1 \\
		ContextEncoder~\cite{pathak2016context} & 14.1 & 20.7 & 21.0 & 19.8 & 15.5 \\
		Adversarial~\cite{donahue2016adversarial} & 17.7 & 24.5 & 31.0 & 29.9 & 28.0 \\
		Colorization~\cite{zhang2016colorful} & 12.5 & 24.5 & 30.4 & 31.5 & 30.3 \\
		Split-Brain~\cite{zhang2017split} & 17.7 & 29.3 & \bf  35.4 & \bf  35.2 & \bf  32.8 \\
		Rep-Cnt\textbf{*}~\cite{noroozi2017representation}  & 18.9 & \bf 30.7 & 33.9& 30.6 &  26.0\\
		Rep-Cnt+$\text{SVMax}''$ (\autoref{eq:rep_cnt_svmax} $\lambda=10$) & \bf 19.4 & 29.3 & 31.7 & 28.9&  24.5 \\
		Rep-Cnt+$\text{SVMax}''$ (\autoref{eq:rep_cnt_svmax} $\lambda=100$) & 19.2 & 29.4 & 31.8 & 29.3 & 25.3 \\
		\bottomrule
	\end{tabular}
	\label{tbl:svmax_rep_cnt}
\end{table}

To integrate SVMax in Rep-Cnt, we replace the problem-specific term with our generic prior as follows
\begin{equation}\label{eq:rep_cnt_svmax}
	L =  \mathbb{D}_{I,T_i} - \lambda \underbrace{s_\mu}_{Generic-prior}
\end{equation}
 where $s_\mu$ is the mean singular value of the mini-batch embeddings $N(I)$. We leverage the unbounded SVMax formulation $(-\lambda s_\mu)$ instead of the bounded SVMax $\left(\lambda \exp\left(\frac{U-s_\mu}{U-L}\right)\right)$ because vector-norms  satisfy the triangle inequality property,~\ie $||x+y||^2 \le ||x||^2 + ||y||^2$. If we normalize the output embedding, the $N(I) = \sum_{i=1}^4{N(T_i)}$ objective becomes infeasible.
 
 To evaluate SVMax quantitatively, we follow Rep-Cnt technical details. We use AlexNet with three fully connected layers. The last fully connected layer provides a feature embedding. Thus, we reduce the layer's dimension from 1000 to 128. This reduces the computational cost of SVMax. We set $m = 10$ in ~\autoref{eq:rep_cnt} as in~\cite{noroozi2017representation}. We pretrain the AlexNet network  using both ~\autoref{eq:rep_cnt} and  ~\autoref{eq:rep_cnt_svmax} as a self-supervision signal. For each pretrained network, we train a linear classifier on top of the frozen CNN layers using ImageNet~\cite{deng2009imagenet}. This evaluation configuration is proposed by~\cite{zhang2016colorful}.
 
 \autoref{tbl:svmax_rep_cnt} presents a quantitative SVMax evaluation in self-supervised learning. We applied SVMax on Rep-Cnt because it is a simple baseline. Rep-Cnt + SVMax does not achieve state-of-the-art results. However, SVMax can be applied on top of various self-supervised pretexts. Through this experiment, we demonstrate how SVMax avoids a model collapse in self-supervised learning. SVMax avoids the trivial solution without the need for a problem-specific repulsion term, an input-reconstruction term, or an adversarial loss term.

\end{document}